\pdfoutput=1

\documentclass[11pt]{article}

\usepackage{acl}

\usepackage{times}
\usepackage{latexsym}

\usepackage[T1]{fontenc}

\usepackage[utf8]{inputenc}

\usepackage{microtype}
\usepackage{hyperref}
\usepackage{comment}
\usepackage{inconsolata}

\usepackage{graphicx}

\usepackage{float}
\usepackage{amssymb}
\usepackage{booktabs}
\usepackage{tabularx}
\usepackage{tablefootnote}

\definecolor{darkgreen}{RGB}{0,100,0}

\newcommand{\mtd}{\texorpdfstring{\texttt{M$\rightarrow$D}}{M->D}}
\newcommand{\dtm}{\texorpdfstring{\texttt{D$\rightarrow$M}}{D->M}}
\newcommand{\catchyname}{DialUp}
\newcommand{\combined}{\texorpdfstring{\texttt{M$\leftrightarrow$D}}{M<->D}}
\newcommand{\mm}{M2M}
\newcommand{\aya}{Aya-23}
\newcommand{\ttt}[1]{\texttt{#1}}

\newcommand{\suffix}[1]{\textcolor{orange}{#1}}
\newcommand{\functionword}[1]{\textcolor{blue}{#1}}
\newcommand{\soundchange}[1]{\textcolor{violet}{#1}}

\title{DialUp! Modeling the Language Continuum by Adapting Models to Dialects and Dialects to Models}

\author{Niyati Bafna\textsuperscript{1}, Emily Chang\textsuperscript{2}, Nathaniel R. Robinson\textsuperscript{1}, \\ \textbf{David R. Mortensen\textsuperscript{3}, Kenton Murray\textsuperscript{1}, David Yarowsky\textsuperscript{1}, and Hale Sirin\textsuperscript{1}} \\
         \textsuperscript{1}Johns Hopkins University, Center for Language and Speech Processing; \\ \textsuperscript{2}University of Virginia; \textsuperscript{3}Language Technologies Institute, Carnegie Mellon University  \\
         \texttt{\{nbafna1,nrobin38\}@jhu.edu}, \texttt{echang22911@gmail.com}
         }

\begin{document}
\maketitle

\begin{abstract}

Most of the world's languages and dialects are low-resource, and lack support in mainstream machine translation (MT) models. 
However, many of them have a closely-related high-resource language (HRL) neighbor, and differ in linguistically regular ways from it. 
This underscores the importance of model robustness to dialectal variation and cross-lingual generalization to the HRL dialect continuum. We present DialUp, consisting of a training-time technique for adapting a pretrained \textbf{m}odel to dialectal \textbf{d}ata (\mtd{}), and an inference-time intervention adapting dialectal \textbf{d}ata to the \textbf{m}odel expertise (\dtm{}). 
\mtd{} induces model robustness to potentially unseen and unknown dialects by exposure to synthetic data exemplifying linguistic mechanisms of dialectal variation, whereas \dtm{} treats dialectal divergence for known target dialects.
These methods show considerable performance gains for several dialects from four language families, and modest gains for two other language families.
We also conduct feature and error analyses, which show that language varieties with low baseline MT performance are more likely to benefit from these approaches.\footnote{\url{https://github.com/niyatibafna/dialup/}}

\end{abstract}

\section{Introduction}
\label{sec:intro}

Recent years have seen advancement in MT quality and language coverage, with models such as M2M100 \citep{fan2021beyond} and NLLB \citep{nllb2022} covering $100$ and $200$ languages respectively, as well as multi-purpose generative language models such as the GPT, BLOOM and Llama \citep{muennighoff2023crosslingual,dubey2024llama} series expanding to dozens of languages. 
These models have displayed cross-lingual generalization capabilities for some unseen or low-resource languages, but they tend to suffer significant performance degradation for most \citep{jiao-etal-2023-chatgpt,robinson-etal-2023-chatgpt,ziems-etal-2023-multi,cahyawijaya2024llms,joshi2024natural}. 

The roughly 7000 languages in the world can be grouped into a few hundred families.
Closely-related languages (CRLs) and dialects within a family largely exhibit continuous and structured differences along phonological, morphological, and lexical dimensions, rather than being discrete monoliths \citep{purschke-2018-dialectcontinuum,bergman-diab-2022-towards}.
Not only is it unfeasible to collect training data for every lect, but language is fluid and dynamic across and between dialect boundaries, calling for principled general approaches to dialectal variation.

\begin{figure}[t]
  \centering
  \includegraphics[width=0.8\columnwidth]{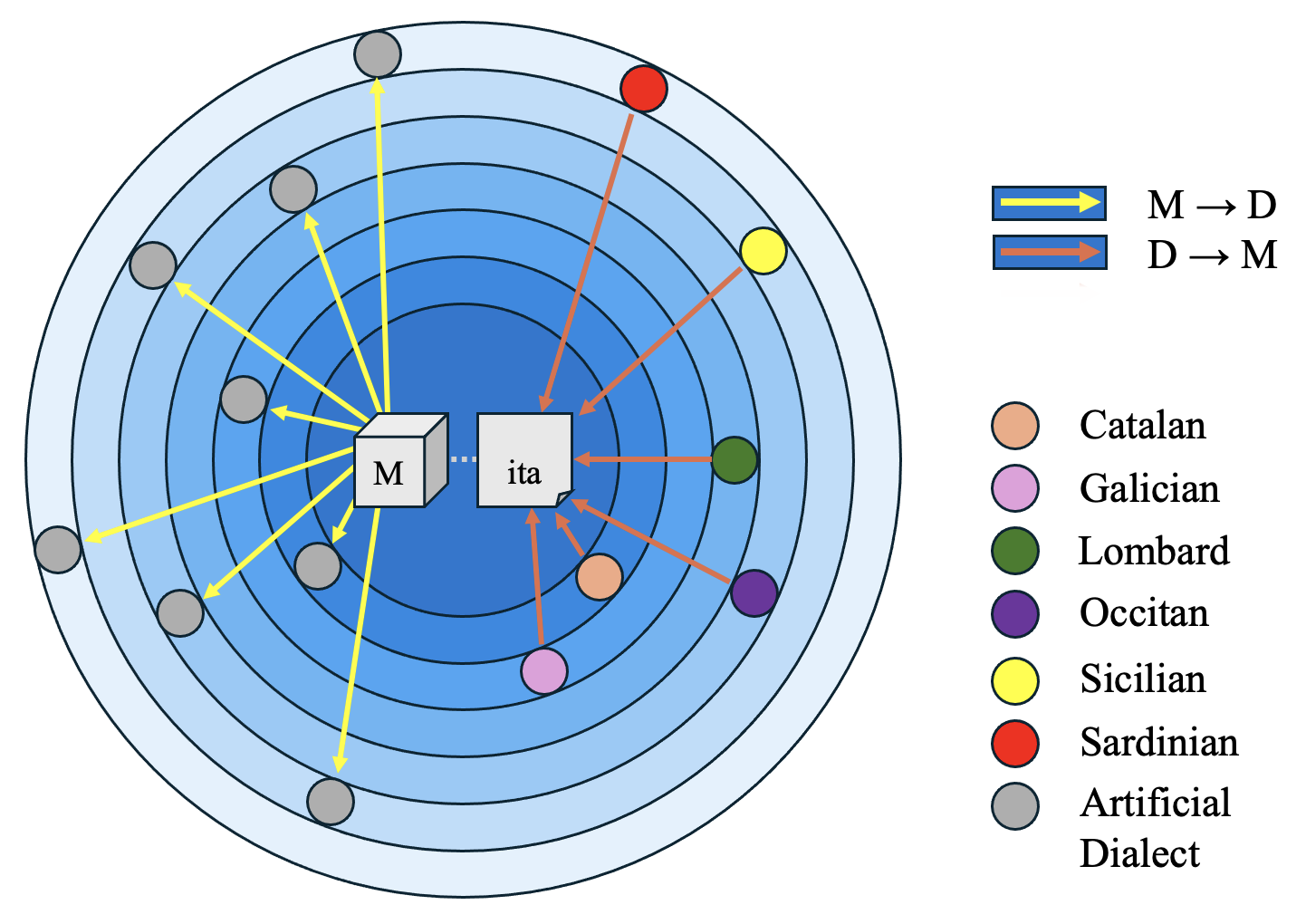}
  \caption{Two paradigms for robustness to dialects on a continuum of distances from an HRL. DialUp involves \mtd{}: training the \textbf{model} on artificial \textcolor{gray}{dialectal variation}, and \dtm{}: bringing dialectal \textbf{data} closer to model expectations (HRL-like input) at inference.}
  \label{fig:first_page}
\end{figure}

Many language families have at least one HRL member that is supported by state-of-the-art models. 
We seek to expand models proficient in HRLs to their lower-resource CRLs, by inducing robustness to unseen varieties across language continua. 
We do this with \catchyname{}: adapting models from an HRL to its language relatives, and adapting those relatives' data inversely towards the HRL model, as in \autoref{fig:first_page}.

We first aim to adapt the model to the language continuum, or \textbf{model-to-data (\mtd) adaptation} (yellow arrows in \autoref{fig:first_page}). 
The goal is robustness to unseen CRLs (both known CRLs absent in pre-training data and yet undocumented CRL varieties), 
using only HRL bitext. 
We do this by simulating dialectal variation patterns over HRL fine-tuning text, teaching the model to generalise to realistic variants such as actual CRL inputs. 
 
Next, we aim to adapt data to the HRL model, or \textbf{data-to-model (\dtm) adaptation} (red arrows in \autoref{fig:first_page}), in the case that a HRL-CRL bilingual lexicon is available or can be induced. 
In this circumstance, we pull CRL data towards the model distribution at inference time by interchanging words for their HRL counterparts. %

\begin{table}
    \centering
    \resizebox{0.47\textwidth}{!}{
    \footnotesize
    \begin{tabular}{r|cccc}
         \hline
         Turkic ["boils"] & \soundchange{k}ayna\textcolor{orange}{r} (tur) & \soundchange{q}ayna\textcolor{orange}{yar} (azj)& \soundchange{g}aýna\textcolor{orange}{r} (Tuk) & \soundchange{qay}na\textcolor{orange}{y} (crh)   \\ 
         \hline
          Romance ["tree"] &  a\soundchange{lbero} (ita) & á\soundchange{rbore} (glg) & a\soundchange{rbre} (oci)  & à\soundchange{rvulu} (scn)  \\ 
          \hline
         Creole ["today"] & \soundchange{j}odi a (hat) & \soundchange{j}ò\soundchange{r}di (lou) & \soundchange{oz}ordi (crs)  &  \soundchange{z}ordi (mfe) \\
         \hline
         
    \end{tabular}
    }
    \caption{Cognates differ in predictable ways, via \soundchange{sound change} patterns, vowel changes, and new \textcolor{orange}{suffixal paradigms.} }
    \label{tab:cognate_table}
\end{table}

Note that cross-lingual generalization can be viewed as a train-test data mismatch problem: we want our models to work on dialectal data which differs in various ways from the training data of the model (HRL data). 
Our two approaches above then represent two broad paradigms for this general problem in machine learning %
(see Figure \ref{fig:first_page}). Model-to-data adaptation has roots in existing cross-lingual transfer approaches, which approximate CRLs using an HRL, relying on linguistic similarities for transfer; however, this does not train the model to handle dialectal divergences. 
\mtd{} innovates on this by synthetically approximating such divergences. 
By approximating language continua, we may avoid pressing questions raised by discrete language paradigms, e.g. which transfer language to use and how much training data to collect \citep{DBLP:journals/corr/abs-1802-07420}.
\dtm{} adaptation has roots in prior approaches to approximate a model's expected domain in data \citep{nie-etal-2023-cross}. 

These two approaches have different advantages. 
\mtd{} does not require any CRL resources or data and does not require partitioning the continuum into discrete dialects. 
\dtm{} is train-free and can be directly applied even to closed-source models. 
These approaches can also be used in tandem, by applying \dtm{} at inference time to a model adapted via \mtd{}. 
We expect these methods to be most beneficial for unseen CRLs and CRLs that have high linguistic overlap with the HRL, i.e. languages that depend on existing HRL representations in the pre-trained model. 
In sum, we contribute:
\begin{itemize}
    \item \catchyname{}, a principled and inexpensive method to induce robustness over language family continua via \mtd{} and \dtm{} adaptation. 
    \item Evaluations of \catchyname{}'s benefits for \ttt{X$\rightarrow$eng} MT with two models for 49 CRLs across six language (sub)families.
    \item Consistent gains via \mtd{} for low-baseline varieties across 4/6 language families (up to mean $+11.4$ BLEU for Romance languages) 
    \item Gains from \dtm{} adaptation over baselines for certain families and CRLs (yielding up to mean $+12$ BLEU for Indic dialects), showing for the first time that adapting dialectal \textit{function words} is more beneficial than adapting content words  
    \item Evidence that \mtd{} and \dtm{} combine advantageously, and provide a recipe for increasing the flexibility of existing MT models to general dialectal variation.
\end{itemize}

\section{\catchyname{}}
\label{sec:method}

\subsection{Model-to-data (\mtd{})}
\label{sec:mtda}

In this paradigm, given an HRL-proficient model and bitext in that HRL, we adapt the model to unseen CRLs.
We generate varied artificial CRLs of the HRL by simulating mechanisms of dialectal variation over the HRL bitext, and fine-tune on the resulting synthetic data to induce model robustness to dialectal variation.

\begin{table*}[!ht]
\centering
    \includegraphics[scale=0.8]{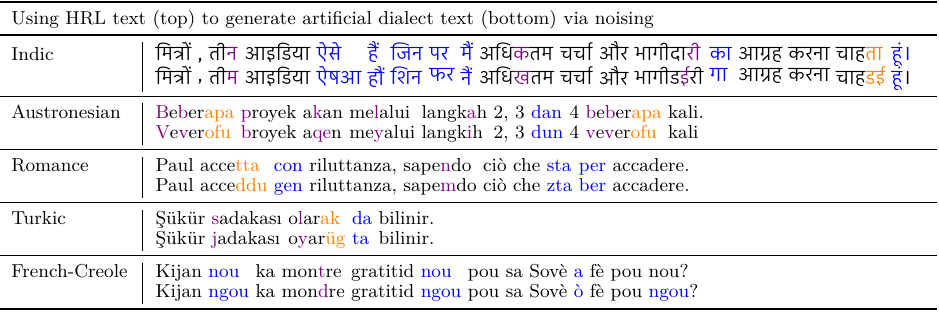}
    \caption{Linguistically-motivated noisers mimic \functionword{function word}, \suffix{suffixal}, and regular \soundchange{sound} changes.}
    \label{tab:noising_examples}
\end{table*}

\paragraph{Artificial language generation}
We employ the approach of \citet{bafna2024evaluating} to simulate language variation along \textbf{p}honological, \textbf{m}orphological, and lexical (\textbf{f}unction word and (non-cognate) \textbf{c}ontent word) dimensions.
Each constituent process operates on the relevant units of the inputs: phonemes, suffixes, and function and content lexemes, and ``noises'' a given unit with probability $\theta^n$, $n$ $\in$ $\{\ttt{p,m,f,c}\}$, replacing all instances of that unit with a plausible alternative in the constructed language (i.e. phones in the original unit randomly replaced with alternate phones of high phonetic proximity, or suffixes swapped with similar-sounding alternatives).\footnote{We provide further details on each noiser in \autoref{app:noising_details}.}
$\theta^n$ serves as a proxy for linguistic distance in each of these dimensions: increase or "dialing up" of $\theta^n$ results in more divergent artificial varieties.
Hence, artificial languages exhibit various kinds of noise, at respective $\theta^n$-dials. 
(See \autoref{tab:noising_examples} for examples of generated artificial language text.) 
Each artificial language is defined as the map of changes made from the HRL. 
We present two alternative ways to distribute these artificial languages: 
\textbf{\mtd{}\ttt{-shell}} generates them
on a single hypersphere at a fixed $\theta^{p,m,f,c}$-radius from the HRL
\footnote{We use $\theta^{p,m,f,c}_{0.05,0.3,0.5,0.001}$, as an average CRL-HRL distance given $\theta^n$ posteriors over several CRL-HRL pairs \citep{bafna2024evaluating}. See \autoref{app:choosing_noising_parameters} for more details.},
while \textbf{\mtd{}\ttt{-cloud}} generates them on multiple hyperspheres around the HRL, populating the hypothesized dialect continuum.\footnote{We use $K=10$ hyperspheres at uniform intervals in each dimension from the HRL and $\theta^{n_i}$ = $\theta^{n}_{max} \cdot \frac{i}{K}$. 
Each hypersphere is therefore used for noising $1/K^{th}$ of the total data.
We set $\theta_{max}$ as $\theta^{p,m,f,c}_{0.07,0.5,0.8,0.001}$, as reasonable maximum CRL-HRL distances in each dimension.
See \autoref{app:choosing_noising_parameters} for more details and tuning experiments for both variants.}
(\autoref{fig:first_page} depicts \mtd{}\ttt{-cloud}; \mtd{}\ttt{-shell} would show all yellow arrows extending to the same blue band.)

\subsection{Data-to-model (\dtm{})}
\label{sec:dtma}
In this paradigm, given CRL-HRL bilingual lexicons, we swap divergent parts of CRL input with known HRL equivalents to pull data towards the model's proficiency distribution. 
Languages from the same family are generally syntactically similar or monotonic \citep{britannica_romance_langs}; hence this switching should largely maintain comprehensible grammatical structure.

We explore three \dtm{} settings: \ttt{func}, \ttt{cont}, and \ttt{all}; in which we swap out only function words,\footnote{determiners, adpositions, auxiliaries, conjunctions, pronouns, and determiners} only content words,\footnote{all other word classes} and both, respectively. 
Function and content word classes differ both in their role in language, with the former crucial for grammaticality and coherence, and the latter providing semantics, as well as in the extent to which they are affected by language change: function words diverge quickly and often opaquely across dialects as compared to less frequent content words \citep{ellis2008dynamics}. It is therefore relevant to isolate the effects of \dtm{} on these word classes. 

Note that isolating these settings requires function word identification in CRL input. 
We achieve this by collecting HRL function words from the Universal Dependencies corpus \citep{nivre2016universal} and annotation projection using the HRL-CRL lexicons. Any word that is not identified as a function word is considered a content word. 

\begin{figure*}
    \centering
    \includegraphics[width=1\linewidth]{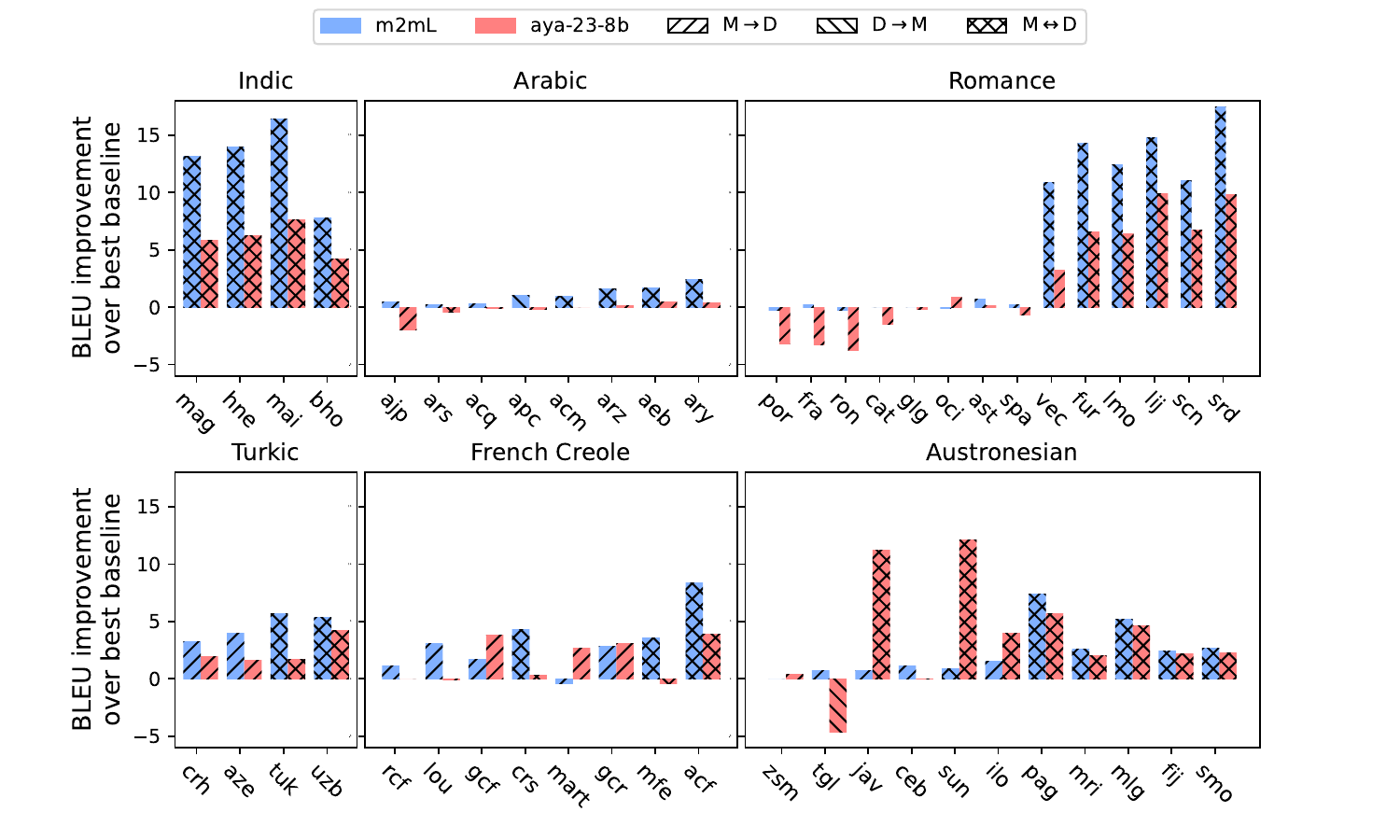}
    \caption{BLEU point improvement of the best \catchyname{} method (\mtd{}, \dtm{}, or \combined{}) over the best baseline (\ttt{off-the-shelf}, \ttt{fthrl}, or \ttt{randaug}). Languages are ordered by their \mm{} \ttt{off-the-shelf} performance. 
    }
    \label{fig:bar_charts}
\end{figure*}

\subsection{\combined{} and baselines}
\label{sec:mtda_dtma}
We also combine 
\mtd{} (\ttt{-cloud} fine-tuning) with \dtm{} (inference-time adaptation) into \combined{}. %

We compare all proposed approaches to three baselines: (1) evaluating the model on CRL$\rightarrow$\ttt{eng} MT without any adaptation 
(\textbf{\ttt{off-the-shelf}}), %
(2) evaluating on CRL$\rightarrow$\ttt{eng} after fine-tuning on ordinary HRL$\rightarrow$\ttt{eng} bitext without any simulated linguistic variation (\textbf{\ttt{fthrl}}), %
and (3) fine-tuning and evaluation after augmenting the HRL$\rightarrow$\ttt{eng} bitext with completely random (i.e. not linguistically motivated or plausible) variations (\textbf{\ttt{randaug}}) %
For the final baseline, we \textbf{r}andomly swap out \textbf{c}haracters in the HRL text with uniformly sampled target same-script characters, and \textbf{w}ords with different words from the source language vocabulary at probability ($\theta^{rc}$, $\theta^{rw}$). 
This is in similar to prior work \citep{belinkov-bisk-2018-synthetic,heigold-etal-2018-how} that introduces random character and word perturbations in the HRL as a method of data augmentation. 
We implement \ttt{shell} and \ttt{cloud} variants of this baseline analogously to \mtd{}, maintain analogous parametrization for a fair comparison\footnote{See \autoref{app:choosing_noising_parameters} for details.}, and only report the better of the two.

\section{Experimental Setup}
\label{sec:exp_setup}

We work with X$\rightarrow$\ttt{eng} MT for 49 languages in six (sub)families.

\paragraph{Languages and Datasets} 

We include the following language groups and designate one HRL in each: Austronesian (with HRL Indonesian), Indic (Hindi), Turkic (Turkish), Romance (Italian), Arabic (Standard Arabic), and French-related Creole languages (Haitian). 
(See \autoref{table:languages} in the Appendix for a full list.) 
We use \texttt{Wikitext} bitext \citep{schwenk-etal-2021-wikimatrix} in the HRLs as training data; we do not use any CRL bitext. All compared methods use the same amount of total HRL bitext ($100K$ sentences).
We included CRLs per language family according to availability of evaluation data in the FloRes-200 \citep{nllb2022} dataset, maintaining a single script within each family. 
We use Kreyòl-MT evaluation sets for Creole CRLs (absent in FloRes) \cite{robinson-etal-2024-kreyol}. 
Our set of CRLs includes languages on a spectrum of relatedness to each HRL, as well as variety of resource levels. 
Some of the CRLs we include, such as French, are high-resource themselves, while most are low-resource.
The Turkic, Austronesian, and Arabic languages we included vary widely in HRL proximity; some have a high degree of mutual intelligibility with the HRL (Azerbaijani, Malay, Saudi Arabic), and some are quite distant (Uzbek, Tagalog, Moroccan Arabic) \citep{glottolog,2014-nouri}.

\paragraph{Bilingual lexicons}

For each CRL-HRL pair, we use a combination of PanLex \citep{kamholz-etal-2014-panlex}, and Swadesh lexicons.\footnote{\url{https://en.wiktionary.org/wiki/Category:Swadesh_lists_by_language}} 
We also added Art Dieli's Sicilian-Italian dictionary \cite{dieli} and Indic dialect lexicons from \citet{bafna-etal-2024-cousin}. However, most of these have little coverage of function words. 
We therefore additionally used (function word-inclusive) lexicons obtained from performing statistical alignment \citep{dyer2013simple} on FloRes \ttt{dev} data.\footnote{See \autoref{app:survey_of_lexicons} for our survey of available lexicons, details and statistics of used lexicons. We also report the type coverage of CRL text for each lexicon, and note that this is generally low for PanLex and Swadesh.}
The Creole CRLs are not included in any of the above lexicon datasets; we used statistically aligned lexicons from JHU Bible translations \citep{mccarthy-etal-2020-johns}, available for \ttt{acf}, \ttt{crs}, and \ttt{mfe}. We therefore only report \dtm{} and \combined{} results for these three Creole CRLs.

\paragraph{Models} 

We employ two MT models in our experiments: one multilingual supervised \ttt{seq2seq} model, M2M-100-1.2B \cite{fan2020englishcentric}; and one multilingual instruction-tuned LLM, \aya{}-8B \cite{aryabumi2024aya}. 
We selected these models because they are highly multilingual (supporting all our selected HRLs) but lack support for enough of our CRLs to warrant legitimate evaluation of adaptation to unseen languages. 
\mm{} only supports 15 of our 49 selected CRLs, while \aya{} supports only 4. 
We selected one \ttt{seq2seq} model and one LLM in order to evaluate our methods in these two settings. 
We curated translation instructions from our bitexts to fine-tune \aya{}. 
(See \autoref{app:aya_prompt} for details.) 
We use LoRA \citep{hu2021loralowrankadaptationlarge} for a single epoch for all fine-tuning processes.

\section{Results}
\label{sec:results}

\begin{table*}[htbp]
	\centering
	\resizebox{1\textwidth}{!}{
	\begin{tabular}{l l | r r r r r r | r | r r r r r r | r}
		\toprule
		& & \multicolumn{7}{c|}{\textbf{\mm}} & \multicolumn{7}{c}{\textbf{\aya}} \\
		\midrule
		   & & AUS (9) & ARA (7) & ROM (6) & TUR (4) & IND (4) & CRE (8) & Mean & AUS (9) & ARA (1) & ROM (5) & TUR (4) & IND (4) & CRE (8) & Mean  \\
\midrule
\textbf{Baselines} & \ttt{\textbf{off-the-shelf}} & 10.8 & 20.2 & 9.6 & 5.0 & 12.6 & 5.9 & 10.7 & \underline{7.6} & \underline{23.5} & 17.1 & 7.9 & \underline{20.4} & 7.3 & 14.0  \\
& \ttt{\textbf{fthrln}} & \underline{+0.3} & \underline{+0.3} & +0.5 & -0.1 & \underline{+0.4} & \underline{+0.9} & \underline{+0.4} & -0.6 & -1.8 & +1.4 & \underline{+1.2} & -1.8 & \underline{+2.3} & \underline{+0.1}  \\
& \ttt{\textbf{ftrandaug}} & +0.2 & +0.2 & \underline{+0.6} & \underline{+0.2} & \underline{+0.4} & \underline{+0.9} & \underline{+0.4} & -0.4 & -2.3 & \underline{+1.9} & +1.1 & -1.8 & +2.1 & \underline{+0.1}  \\
\midrule
\mtd & \ttt{\textbf{-shell}} & \textbf{\underline{+1.9}} & \underline{+1.0} & \underline{+11.5} & \underline{+2.5} & +5.0 & \underline{+2.7} & \underline{+4.1} & +2.1 & \textbf{\underline{+0.4}} & \underline{+7.9} & +2.9 & \underline{+2.8} & +3.0 & \underline{+3.2}  \\
& \ttt{\textbf{-cloud}} & +1.3 & +0.9 & +9.1 & +1.6 & \underline{+5.4} & \underline{+2.7} & +3.5 & \underline{+2.2} & +0.1 & +7.7 & \textbf{\underline{+3.2}} & +2.7 & \textbf{\underline{+3.5}} & \underline{+3.2}  \\
\midrule
\dtm & \ttt{\textbf{-cont}} & -0.4 & -2.2 & +1.6 & +1.4 & +0.3 & +2.5 & +0.5 & +0.8 & -6.2 & -1.0 & -0.6 & -1.1 & -3.8 & -1.9  \\
& \ttt{\textbf{-func}} & -0.1 & \underline{+0.8} & +8.0 & +1.0 & \underline{+12.0} & +3.7 & +4.3 & +2.4 & \underline{-1.8} & \underline{+4.7} & \underline{+0.3} & \underline{+5.6} & \underline{-3.3} & \underline{+1.3}  \\
& \ttt{\textbf{-all}} & \underline{+0.0} & -1.4 & \underline{+9.8} & \underline{+3.0} & +11.4 & \underline{+5.6} & \underline{+4.7} & \underline{+3.7} & -6.9 & +3.1 & +0.0 & +4.4 & -3.5 & +0.2  \\
\midrule
\combined & \ttt{\textbf{-cloud-cont}} & \underline{+1.4} & -1.5 & +9.0 & +2.6 & +5.1 & +3.7 & +3.4 & +3.3 & -4.3 & +5.3 & +1.8 & +1.6 & +2.6 & +1.7  \\
& \ttt{\textbf{-cloud-func}} & +1.1 & \textbf{\underline{+1.3}} & \textbf{\underline{+14.1}} & +2.3 & \textbf{\underline{+13.0}} & +4.6 & \textbf{\underline{+6.1}} & +3.8 & \underline{-0.8} & \textbf{\underline{+9.8}} & \underline{+3.1} & \textbf{\underline{+6.0}} & \underline{+2.8} & \textbf{\underline{+4.1}}  \\
& \ttt{\textbf{-cloud-all}} & +0.9 & -1.2 & +13.1 & \textbf{\underline{+3.9}} & +12.2 & \textbf{\underline{+6.1}} & +5.8 & \textbf{\underline{+5.1}} & -4.5 & +7.4 & +2.1 & +4.6 & +2.5 & +2.9  \\
			 \bottomrule
	 \end{tabular}}
	\caption{BLEU score performance gains relative to \ttt{off-the-shelf}, for low-performing CRLs (\ttt{off-the-shelf} score < 25), averaged by language family. \# of such CRLs per family provided in parentheses. The overall best score is \textbf{bolded} and the best score in each paradigm is \underline{underlined}. Language families are abbreviated with first three letters (e.g. Indic (IND)). \mtd{} and \dtm{} both outperform best baselines; \combined{} is generally the best.}
	\label{bleu_langfam_low_baseline}
\end{table*}

Results across all languages are in \autoref{fig:bar_charts}.
Our approaches give gains across the board, with varied trends by language family. 
Mean gains of low-performing languages ($<25$ BLEU for \ttt{off-the-shelf}) per method and language family are in \autoref{bleu_langfam_low_baseline}. 
See \autoref{app:more_results} for detailed results and COMET \citep{rei2020cometneuralframeworkmt} scores. We found that these show trends consistent with BLEU.

We observe that for both models, languages with poorer baseline scores, which tend to be the low-performing varieties we aim to assist in this work, benefit more from \catchyname{}, while languages with better baselines show small or negative gains. 
This trend is visible in \autoref{fig:bar_charts}, where languages are ordered by their \ttt{off-the-shelf} BLEU score, and especially pronounced for Romance languages.
We also see that \mm{} typically benefits more. This may be because \aya{} was likely exposed to some CRLs in pre-training, despite not being trained explicitly on them, given the heterogenity and poor documentation of LLM pre-training sets. 
Two exceptions to this trend are Javanese (\ttt{jav}) and Sundanese (\ttt{sun}) (gaining $+13.3$ BLEU with \aya{}, and little with \mm{}), which \aya{} translates poorly off-the-shelf, and which \mm{} explicitly supports in pre-training. 
Proximity to the HRL also appears to play a role: CRLs distant from their HRL, such as Samoan and Tagalog, as well as those extremely close to the HRL, such as Malay (\ttt{zsm}), Azerbaijani (\ttt{aze}), and Saudi Arabic (\ttt{ars}), appear to benefit little from \catchyname{}.

\begin{table*}[!ht]
    \centering
    \includegraphics[width=1\linewidth]{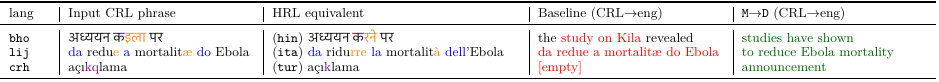}
    \caption{\mtd{} improves \mm{} on input with cognate processes such as \functionword{function words}, \suffix{new suffixes}, and \soundchange{phonological/orthographic} change. The baseline often transliterates the input or leaves it as is.}
    \label{tab:phrases_noising_examples}
\end{table*}

\textbf{\mtd{}} gives mean gains over best baselines for all language families, demonstrating the efficacy of linguistically-motivated (as opposed to random) synthetic data augmentation. \autoref{tab:phrases_noising_examples} shows examples of improvements on cognate words, including new inflections and function words. We often see that while \ttt{off-the-shelf} models transliterated them or left them as are instead of translating, \mtd{}-tuned models decode them correctly.\footnote{See full-length examples in \autoref{app:more_examples}.}
\textbf{\dtm{}} introduces gains for many low-performing varieties---up to $+11.4$ BLEU for \aya{} on Sundanese and $+13.7$ BLEU for \mm{} on Chhattisgarhi (\ttt{hne})---where replacing dialectal words with HRL equivalents is helpful. 
However, it is also damaging for several, especially high-performing CRLs like French (\ttt{fra}), Spanish (\ttt{spa}), and Tagalog (\ttt{tgl}). 
Conceivably, if the base model is already proficient in a CRL, \dtm{} code-switching introduces counterproductive unnaturalness. 
Since \dtm{} is an inference-time method, practitioners may activate or deactivate it according to language needs. 
See \autoref{tab:phrases_denoising_examples} for examples of \dtm\ttt{-func} input and output, demonstrating the brittleness of baseline models on dialectal function words, as well as the impact of treating them.

\begin{figure}[!ht]
    \centering
    \includegraphics[width=\linewidth]{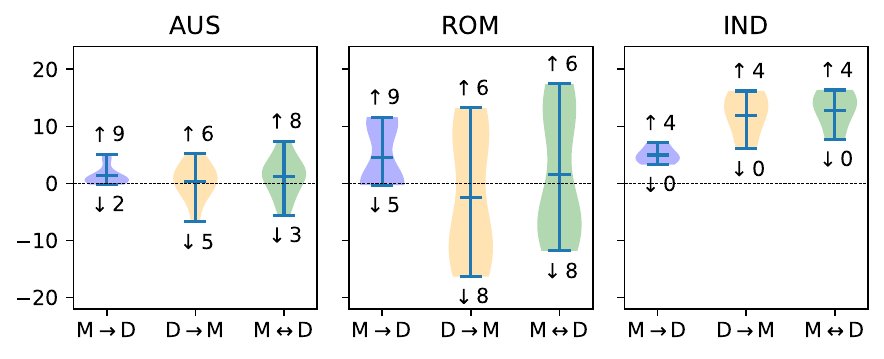}
    \caption{BLEU score improvements over the best baseline with \mm{} for three language families. $\uparrow$ and $\downarrow$: \# CRLs with positive/negative gains. \mtd{} gives more consistent positive gains. 
    }
    \label{fig:results_variance}
\end{figure}

So which paradigm performs better? 
That depends on the model and the language. 
See \autoref{fig:results_variance} for a distribution of score improvements for two families, and number of wins over the best baseline.  
Most families behave similarly to Austronesian and Romance (see \autoref{app:variance} for similar plots for all languages/models).
In general \mtd{} gives lower-variance gains, with wins over the baseline for most CRLs. 
\dtm{} often shows similar or higher mean gains but with higher variance, and consistently fewer wins. 
Indic languages show a different distribution, with a unanimous preference for \dtm{}. 
Notably, \combined{} generally performs best across languages. 
(See \autoref{bleu_langfam_low_baseline}.)
Holistically, results suggest that \textbf{\mtd{} provides gains with low-risk consistency, and that \dtm{} provides significant benefits for a select set of CRLs.
}

\begin{table*}[!ht]
    \centering
    \includegraphics[width=\linewidth]{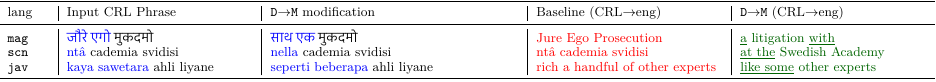}
    \caption{\dtm{} improves \mm{} output by replacing \functionword{function words} in the input (\underline{underlined} in output) with HRL equivalents.}
    \label{tab:phrases_denoising_examples}
\end{table*}

We note that \dtm\ttt{-func} consistently outperforms \dtm\ttt{-cont}, and is often better or close to \dtm\ttt{-all}.
This highlights the importance of treating dialectal function words in enabling model comprehension, and the utility of collecting CRL-HRL function word maps for low-resource languages.
This insight is particularly convenient given that function words form a small closed class and are much easier to collect comprehensive lexicons for than open-class content words; they are also more likely to be accurately aligned with statistical alignment than relatively rarer content words. 

In fact, exchanging content words is largely unhelpful across the board, including from large high-quality lexicons like Art Dieli for Sicilian (\ttt{scn}).
We attribute this to the higher degree of context dependence in handling content words; synonymy, word sense variation, naturalness, and lexicon noise likely contribute to this problem.

See \autoref{app:exp_variants} for our additional \mtd{} experiments in using multiple HRLs per family, training with more data or for epochs, different source datasets, and more aggressive noising. These mildly improve or maintain current trends. We also show that switching CRL words directly into English instead of the HRL for \dtm{} degrades performance, presumably due to unnaturalness of CRL-eng (as compared to CRL-HRL)code-mixing.

\section{Discussion}
\label{sec:discussion}

\textbf{When does \mtd{} help?} 
To test our hypotheses that baseline performance and proximity to the HRL interact meaningfully with score improvement, and to understand which circumstances render \catchyname{} more or less effective, we analyze how language features correlate with gains. 
We computed Spearman's $\rho$ coefficients between BLEU improvement from \mtd{} over the \ttt{off-the-shelf} baseline and features indicating both baseline CRL support (\ttt{off-the-shelf} BLEU score, whether the model explicitly supports the CRL, and number of CRL Wikipedia pages\footnote{per \citet{robinson-etal-2023-chatgpt}}) and HRL-CRL relatedness (character F-score \citep{popovic-2015-chrf} between HRL and CRL FloRes dev sets and average token-count ratio in CRL dev lines to HRL dev lines when using the model's HRL tokenizer\footnote{In the case of the Creole CRLs, which lack FloRes dev sets, we used JHU Bible translations, available for \ttt{acf}, \ttt{crs}, and \ttt{mfe}.}). 
For both Aya and M2M the only such feature that correlated significantly (p$<$0.01) was \ttt{off-the-shelf} BLEU, with $\rho=-0.52$ and $-0.45$, respectively---indicating \textbf{a moderate-to-strong negative correlation between baseline performance and improvement from \mtd{}}, as hypothesized in \autoref{sec:results}. 

\begin{figure}
    \centering
    \includegraphics[width=\linewidth]{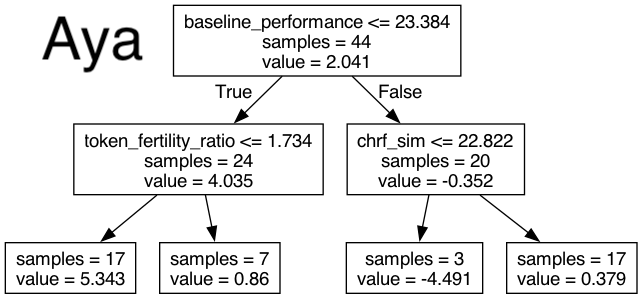}
    \includegraphics[width=\linewidth]{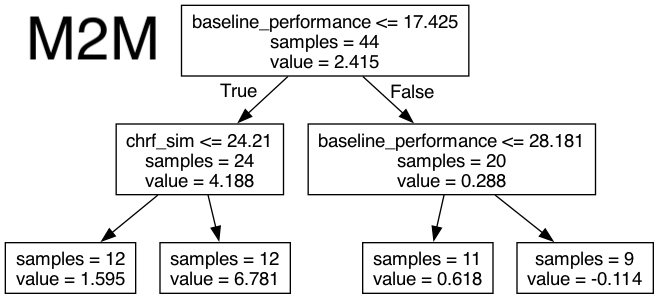}
    \caption{Decision trees indicate that the languages benefiting most from adaptation are low-baseline languages with less than 1.75 times HRL token fertility for Aya, and low-baseline languages with more than 24.2 chrF proximity to the HRL.}
    \label{fig:trees}
\end{figure}

Further analysis also suggests that baseline score is more predictive of \mtd{} success than other features; this is confirmed by random forests fitted over the same features for both models (see decision trees in \autoref{fig:trees}), as well as feature weights from a linear regressor. (See \autoref{feat_imps}.)

\begin{figure}
    \centering
    \includegraphics[width=.49\linewidth]{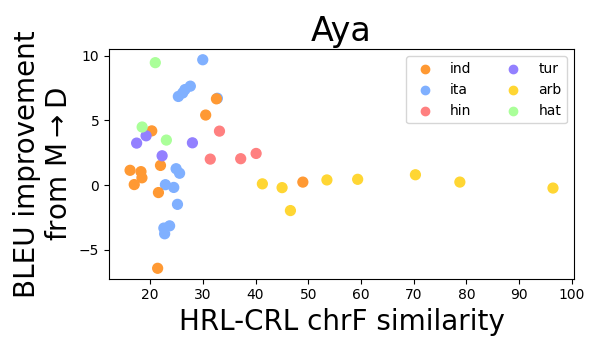}
    \includegraphics[width=.49\linewidth]{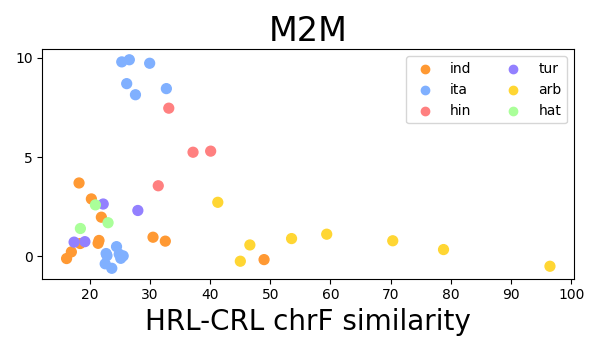}
    \caption{CRL-HRL proximity, measured as character F-score between dev sets, plotted against BLEU increase from \mtd{}, suggests a peak from 20-30 chrF. Arabic varieties are outliers because the FloRes dialectal Arabic sets are unusually close to standard Arabic.
    }
    \label{fig:dots}
\end{figure}

Measurements of HRL-CRL relatedness had the second-highest forest feature importance for both models. Though our measurements of HRL-CRL relatedness do not correlate with BLEU score improvement, we suspect there may be a non-linear relationship between dev-set character-F score and method effectiveness. 
When plotted in \autoref{fig:dots}, the points suggest the contour of a downward-facing parabola, indicating that \mtd{} is most effective for CRLs that are not too close but not too far from the HRL, as we hypothesized. This has an intuitive interpretation: CRLs that are too close may either already be well-performing, or may not display enough dialectal divergence to benefit from DialUp, whereas CRLs that are too far may have a higher amount of non-cognate divergences from the HRL that DialUp does not help with.
However, note that language family is possibly confounded with character-F score, with only Austronesian CRLs covering a wide range on our observed contour; it would be difficult to disentangle the effect of language family without many more languages.

\paragraph{Setting $\theta^{p,m,f}$-dials}
\label{sec:noise_param_search}

\begin{figure}
    \centering
    \includegraphics[width=0.8\linewidth]{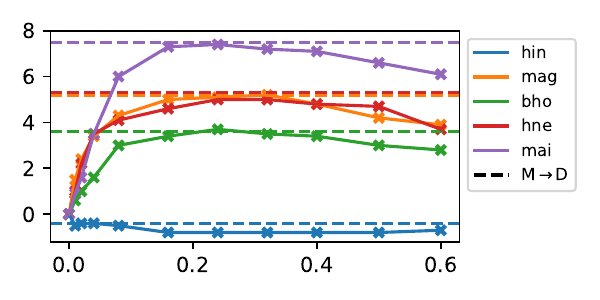}
    \caption{Gains in BLEU points for different values of $\theta^p$ (1-dimensional noiser) for Indic languages, with dotted lines showing the performance of \mtd{}\ttt{-cloud}, using the the 3-dimensional noiser $\theta^{p,m,f}$ with default parameters. Tuning only $\theta^p$ for Indic is competitive with \mtd{}\ttt{-cloud}.}
    \label{fig:noising_param_search}
\end{figure}

Although \mtd{} uses phonological, morphological, and function word noising to model cognate processes in CRLs, the latter two are specialized versions of the phonological noiser, applying it at a fixed high dial to HRL suffixes/function words (\autoref{app:noising_details}).
We show that this $3-$dimensional noiser can be approximated solely by the phonological noiser at a suitable $\theta^p$-dial.
Setting all other $\theta^n=0$, we sweep over $\theta^p$ for Indic (\autoref{fig:noising_param_search}). 
We observe that with an optimal $\theta^p$, the phonological noiser achieves competitive gains as \mtd{}\ttt{-cloud}, which uses all noisers.
Our default \mtd{} $\theta$-dials are eminently reasonable, yielding the best possible performance achievable by the above procedure; however, these defaults may be more suited to particular language families than others. Given that tuning $\theta^{p,m,f}$ in combination is intractable, we therefore recommend simply tuning $\theta^p$ for a given language family as a good proxy, along with trying our defaults.
The above results indicate that \textbf{model gains from \mtd{} largely arise from learning to decode sound changes in dialects}, regardless of their placement in words.

\paragraph{\mtd{} error analysis}
Arabic varieties gain less than other groups from \mtd{} and \combined{}. 
This may indicate that our \mtd{} method does not approximate the Arabic linguistic variation well. 
Arabic varieties do exhibit phonological differences (which \mtd{} simulates). 
However, these are often not expressed in the orthography, making lexical, syntactic, and register differences more prominent differentiators in text. 
We experimented simulating lexical choice variation in CRLs using HRL WordNets as part of our artificial language generation method for 5 language families; however, this did not give us significant gains (\autoref{app:exp_variants}). 
\catchyname{} in its current form may not be well suited for Arabic. 
Additionally, we note that FloRes sets are extremely close to standard Arabic, with a mean dev "dialectness" score
\cite{keleg-etal-2023-aldi} of only 28\%.\footnote{Comparatively, MADAR-26 dev sets \cite{bouamor-etal-2018-madar} for the same varieties achieve 59\%. We initially attempted to use this dataset, but the genre mismatch with train data rendered both baseline and challenger scores too low.}
This may also limit the scope of improvements from \catchyname{} methods on this test set.

French-related Creole languages, which also had low baseline scores and small \mtd{} gains, may have also suffered from genre mismatch (given their heterogenous test sets \cite{robinson-etal-2024-kreyol}). 
We initially hypothesized that their poor performance may be due to the HRL's (Haitian) poor baseline performance with \aya{}. 
We attempted curriculum training with fine-tuning first on Haitian bitext and then with \mtd{}; however this did not materially help.
It is still possible that given Haitian's low-resource nature (and its extremely low-resource CRLs), training token frequencies are too low to risk obfuscating them with \catchyname{} noise. 
We hope future work will find more concrete conclusions on this trend. 

\paragraph{\dtm{} error analysis}

We investigate why \dtm{}\ttt{-func} degrades the performance of Romance and Austronesian HRLs, while benefiting several CRLs in these families and Indic. 
We perform a manual evaluation of 100 words from our function word lexicons for 4 languages from these families (\autoref{tab:manual_eval_lexicons})\footnote{See \autoref{app:manual_evaluation} for details.} to characterize the extent of noise in the \dtm{}\ttt{-func} pipeline: a CRL word may be misidentified as a function word (mean function word identification accuracy: 83.6\%), or it may be mistranslated (mean function word translation accuracy: 56.5\% , general translation accuracy: 59.7\%). 
The introduced noise is naturally particularly damaging when the baseline translation is good (as in for many high-resource CRLs). 
In fact, we find that \dtm{} swaps hurt for high-resource CRLs even with ideal-case accurate swaps from clean lexicons.
On the other hand, for a number of CRLs such as Maithili (mai), we observe that the benefits of making key function words swaps outweighs the negative impact of this noise; we find that even noisy automatically collected lexicons contribute to \dtm{} improvements in some languages (e.g. excluding relatively noisy \citet{bafna-etal-2024-cousin} lexicons results in a drop of $9$ BLEU for Maithili with \mm).
\dtm{} benefits also depend on the model and HRL: in the case of Haitian, \dtm{} performs well with \mm{} (yielding +7 BLEU for \ttt{acf}), but degrades baseline performance for \aya{}. This can be explained by the fact that \aya{} has <10 BLEU performance on Haitian itself, making it unhelpful to switch other CRL words into Haitian; on the other hand, \mm{} baseline is strong for Haitian but weak for related CRLs, rendering \dtm{} a useful strategy. 
\textbf{We therefore recommend \dtm{} for low-baseline CRLs, with a model that performs well on the language family HRL.
}

Note that \dtm{} shows small gains for Arabic and Turkic CRLs.
This is likely caused by a scarcity of word exchanges, which \dtm{} relies on for increasing CRL input comprehensibility.
In fact, \dtm{}\ttt{-func} results in only 16.3\% and 15\% of words being swapped for Arabic and Turkic, respectively, compared to 43.1\% for Austronesian, 38.5\% for Romance, and 33.4\% for Indic. (See \autoref{app:func_word_collection}). This is potentially a result of Arabic and Turkic languages' comparatively complex morphology, in addition to above noted issues with Arabic test sets. 
Grammatical morphemes in these languages are frequently affixes, clitics, or attached to affixes and seldom occur on their own; these will be missed by our whitespace-reliant word-switching technique.
Future work can investigate integrating morphological analysis technologies when available into our methods.

\section{Prior Work}
\label{sec:related_work}

\catchyname{} has theoretical bases in a large body of prior work of noise pattern induction for robustness. Similar to \textbf{\mtd{}} previous researchers have developed methods of injecting noise into training data, including orthographic variants or typos \cite{belinkov-bisk-2018-synthetic,heigold-etal-2018-how}, grammatical errors \cite{anastasopoulos-etal-2019-neural}, learned noise types \cite{brahma-etal-2023-selectnoise}, and bilingual lexicon-induced code-mixing \cite{jones-etal-2023-gatitos,xia2019generalized}. These approaches require monolingual data or bilingual lexicons for from-scratch training. \mtd{} innovates by inducing general dialectal robustness in a pre-trained model to unseen dialects without knowledge or resources in target varieties. \textbf{\dtm{}} also draws from prior work, namely input processing techniques for LLM in-context learning via CRL-English lexicons, morphological analyzers, and grammar references \citep{zhang-etal-2024-teaching,ghazvininejad2023dictionary,zhang-etal-2024-hire,tanzer-etal-2024-benchmark,dimakis-etal-2024-dictionary}. \dtm{} is likewise an inference-time intervention, but it is applicable to both traditional MT models and LLMs. Notably, our exploration of \dtm{} is the first to validate the insight that model comprehensibility of often-ignored function words is beneficial, and to show considerable gains using small, automatically collected lexicons. 

\section{Conclusion}
We present DialUp, a general recipe for expanding pretrained model coverage from its high-resource training languages to their dialect continua. This involves a finetuning-based technique involving artificial dialect generation using only HRL data, as well as an inference-time technique for rendering dialectal input more comprehensible to the model. DialUp shows large gains on several low-resource languages, and introduces a promising paradigm for making NLP flexible to potentially unseen and undocumented dialectal variation.

\section*{Limitations}
\label{sec:limitations}

\paragraph{Language-family-dependent gains} The benefits of \mtd{} and \dtm{} clearly vary by language family: Indic and Romance benefit much more than other families, and Arabic in particular benefits minimally. 
\mtd{} relies on approximating realistic synthetic dialects of the HRL by simulating various mechanisms of language variation. 
While these are broadly similar across language families and can be grouped into phonological, morphological, and lexical processes, which our artificial language generation method seeks to model, each language family naturally varies in typology, intra-family linguistic diversity, and the kinds of relationship that its CRLs have with each other and with the HRL.
For example, certain variants may differ from the HRL in their extent of lexical influence from a third (colonial) language, as in the cases of some Arabic and Turkic languages.
Certain typologies or scripts may also be more or less amenable to our methods; e.g. we observe that suffix stacking in Turkic languages is challenging for the morphological noiser, which performs suffix identification in a manner most suited for fusional or morphologically non-complex languages \cite{ismayilzada2024morph}.
Finally, some language families, such as Arabic, may exhibit syntactic variation, which is not modeled at all by our current method. 
Our current artificial language generation method does not cater to these family-specific characteristics, which may contribute to its poorer performance on these families.
We leave it to future work to refine the current catch-all technique of artificial language generation for specific families.

\paragraph{Different-script CRLs} \mtd{} does not support artificial language generation across scripts, i.e. it cannot perturb an HRL in one script into an artificial language in a different script. This is relevant for language families with more than one script: for example, several Turkic low-resource languages use the Cyrillic script, and cannot be supported with our current methods. Future work can look into transliteration-based approaches to handle this issue. 

\paragraph{Models} We provide results on one representative model of the traditional NMT as well as LLM paradigm, choosing an appropriate model with respect to its training languages given our goal of handling unseen and unknown target dialects, and available evaluation data. Naturally, our method should be tested on a variety of models from each paradigm to gauge its usefulness as a general data augmentation technique for achieving dialectal robustness.

\paragraph{Availability of bilingual lexicons} Our best-performing \dtm{}\ttt{-func} approach depends on the availability of CRL-HRL function word mappings. These are much easier to collect than bilingual lexicons given that they are a small class of closed class words, and we show that using even noisy alignments from a small amount of bitext is effective for this purpose: however, this may still be unfeasible for particularly low-resource languages.

\paragraph{Long way to go} While our methods show considerable improvements on cognates for CRLs (examples in \autoref{tab:phrases_noising_examples}), and improved scores using automated MT evaluation metrics such as BLEU and COMET, the improved translation may still not be usable or coherent. We provide examples of this in \autoref{tab:mtd_examples}.

\section*{Ethics Statement}

Our work seeks to expand the benefits of mainstream NLP (specifically, machine translation) in standard high-resource varieties of languages to their diverse, dialectal, colloquial, and/or low-resource neighbours. 
Our work introduces general techniques towards this goal; however, any particular language community may have specific and different needs, potentially not served by mainstream NLP, which our work does not take into account \citep{bird-2020-decolonising}.
While we attempt to use minimal resources in low-resource languages to achieve our goal in order to make our methods resource-light and widely applicable, using only HRL data in training risks reinforcing a culturally mainstream view of the world in our models.
This may be particularly problematic for continua where dialect-speaking communities diverge in political or other views from communities speaking the standard variety.

\bibliography{acl}

\appendix

\newpage
\mbox{}
\newpage
\section{Languages}
\label{app:languages}

\begin{table*}[t]
\centering
\resizebox{\textwidth}{!}{
\footnotesize
\begin{tabular}{llllllll}
\hline
\textbf{Language} & \textbf{Family} & \textbf{Subfamily} & \textbf{ISO code} & \textbf{FloRes/NLLB} & \textbf{In M2M?} & \textbf{In Aya-23?}  & \textbf{Resource Level} \\
\hline
Indonesian & Austronesian & Malayo-Polynesian & ind & ind\_Latn & id & ind  & HRL \\
\hline
Javanese & Austronesian &Malayo-Polynesian &jav &jav\_Latn & -& - &  <200K \\
Sundanese & Austronesian & Malayo-Polynesian&sun & sun\_Latn &- &  -& <100K \\
Samoan & Austronesian & Malayo-Polynesian, Polynesian &smo & smo\_Latn& - & -& <10K \\
Maori & Austronesian & Eastern Polynesian &mri & mri\_Latn &-  & -& <50K \\
Cebuano & Austronesian & Malayo-Polynesian &ceb & ceb\_Latn& - & -& <12M \\
Standard Malay & Austronesian & Malayo-Polynesian &zsm  & zsm\_Latn &- &  -& <2M \\
Tagalog & Austronesian & Malayo-Polynesian, Philippine &tgl &tgl\_Latn & -  &- & <500K \\
Ilocano & Austronesian & Malayo-Polynesian, Philippine &ilo & ilo\_Latn& - &- & <100K \\
Fijian & Austronesian &Eastern Malayo-Polynesian &fij &fij\_Latn &-  &- & <10K \\
Plateau Malagasy & Austronesian & Malayo-Polynesian, Western Indonesian &plt & plt\_Latn & -  &- & <500K\\
Pangasinan & Austronesian & Malayo-Polynesian, Philippine&pag & pag\_Latn &- & - &  <10K \\
\hline
Arabic (MSA) & Arabic & Modern Standard Arabic &arb & arb\_Arab & ar & ara &   HRL \\ 
\hline
Mesopotamian Arabic & Arabic & Mesopotamian Arabic & acm & acm\_Arab&- & - & <1K\\
Ta'izzi-Adeni Arabic & Arabic & Southern Yemeni  &acq & acq\_Arab& - & -&  <1K\\
Tunisian Arabic & Arabic & Maghrebi &aeb & aeb\_Arab &- & - &  <1K \\
South Levantine Arabic & Arabic & Levantine Arabic & ajp & ajp\_Arab& & - &  <1K\\
North Levantine Arabic & Arabic & Levantine Arabic& apc & apc\_Arab& -& -&  <1K \\
Najdi Arabic & Arabic& Peninsular Arabic & ars & ars\_Arab &- & -&   <1K\\
Moroccan Arabic & Arabic& Maghrebi & ary & ary\_Arab&- &- &  <100K \\
Egyptian Arabic & Arabic & Egyptian Arabic& arz & arz\_Arab & -& -&   <3M \\
\hline
Italian & Romance & Italo-Western & ita & ita\_Latn& it & ita &   HRL\\
\hline
Spanish & Romance & Italo-Western &spa & spa\_Latn& es & spa&  HRL  \\
French & Romance & Italo-Western &fra &fra\_Latn & fr & fra &  HRL \\
Portuguese & Romance &Italo-Western & por & por\_Latn& pt & por &  HRL \\
Romanian & Romance & Eastern Romance &ron & ron\_Latn& ro & ron&   <3M \\
Lombard & Romance & Italo-Western&lmo & lmo\_Latn &-  &- &  <200K \\
Asturian & Romance & Italo-Western&ast & ast\_Latn&ast &- &  <500K \\
Galician & Romance &Italo-Western &glg & glg\_Latn& gl& - & <500K \\
Venetian & Romance & Italo-Western &vec &vec\_Latn &- & - &  <200K\\
Catalan & Romance & Italo-Western&cat & cat\_Latn & ca&-  & <2M \\
Sicilian & Romance & Italo-Western & scn & scn\_Latn & -&-  & <100K\\
Sardinian & Romance & Southern Romance &srd & srd\_Latn& -& - & <50K\\
Friulian & Romance & Italo-Western&fur & fur\_Latn& -& - &  <10K \\
Ligurian & Romance & Italo-Western &lij & lij\_Latn&- &-  &  <50K\\
Occitan & Romance & Italo-Western &oci & oci\_Latn & oc& -&  <200K \\
\hline
Turkish & Turkic & Oghuz &tur & tur\_Latn & tr & tur & HRL \\
\hline
Azerbaijani & Turkic & Oghuz &  azj & azj\_Latn& az & - & <1M \\
Crimean Tatar & Turkic & Kipchak &crh & crh\_Latn & -&-  & <50K \\
Turkmen & Turkic & Oghuz &tuk & tuk\_Latn &- &- & <50K \\
Uzbek & Turkic & Karluk & uzb & uzn\_Latn & uz & -  & <1M \\
\hline
Hindi & Indic & Western  &hin & hin\_Deva & - &- &  HRL\\
\hline
Awadhi & Indic & Eastern  & awa & awa\_Deva&- & - &  CRL\\
Bhojpuri & Indic & Eastern & bho & bho\_Deva&- &-  &  <100K \\
Chhattisgarhi & Indic & Eastern &hne & hne\_Deva& - &- &  <2M \\
Magahi & Indic & Eastern & mag & mag\_Deva& - &- &  <2M \\
Maithili & Indic & Eastern &mai & mag\_Deva& -& - &   <50K \\

\hline
Haitian & French-related Creole & Haitian Creole &hat & hat\_Latn& ht & hat& HRL\\
\hline
Guadeloupean & French-related Creole& Antillean Creole &gcf & -&- &- &   <1K \\
Martinican & French-related Creole & Antillean Creole& gcf &- & -&- &   <1K \\
Saint Lucian Patois & French-related Creole &Antillean Creole &acf &- &- &- &   <1K \\
French Guianese & French-related Creole & French Guianese Creole& gcr & -&- &- & <1K \\
Louisiana Creole & French-related Creole & Louisiana Creole& lou & -&- &- &   <1K \\
Mauritian & French-related Creole & Bourbonnais Creoles &mfe &- &- & -&  <1K \\
Réunion Creole & French-related Creole & Bourbonnais Creoles &lmo &- &- & -&   <1K \\
Seychellois & French-related Creole & Bourbonnais Creoles  &crs & - & - & - &  <1K \\
\hline

\hline

\end{tabular}
}
\centering

\caption{Languages studied organized by language family. We also report the languages supported by the models used, as well resource-level. We report the number of Wikipedia articles for a language as a proxy for its resource level.}
\label{table:languages}
\end{table*}

See \autoref{table:languages}.

\section{Details of artificial language generation}
\label{app:noising_details}

In \autoref{sec:method}, we introduce the artificial language generation method for \mtd{}, which consists of phonological, morphological, function word, and content word ``noising'', or corruption, taken from \citet{bafna2024evaluating}.
\begin{itemize}
    \item The \textbf{phonological} noiser changes phonemes in a given left and right phonological context to phonetically nearby targets (e.g. \ttt{t$\rightarrow$d}, which differ only in voicing). We use script-to-IPA mappings and vice versa to first convert our input into IPA. We define plausible target phoneme sets for each phoneme based on its phonetic characteristics. The noised phoneme is then converted back into the input script. 
    \item The \textbf{morphological} noiser targets suffixes in the HRL. Suffixes are identified heuristically, collected by frequency over an HRL corpus. A given suffix is noised by applying a fixed high amount of phonological noise over it ($\theta^p$=0.8), and changed globally over all occurrences to its noised version.
    \item Similarly, the \textbf{function word} noiser targets function words in the HRL. Function word identification is done in the following way: we curate a set of POS tags for closed-class words, including determiners, pronouns, conjunctions, auxiliaries, and adpositions. We then use the tagged Universal Dependencies corpus \citep{nivre2016universal} to identify all words in a given language whose most frequent tag is one of these. Since function words are a relatively small set, this procedure should yield coverage even over a small corpus. A given function word is noised by applying a fixed high amount of phonological noise over it ($\theta^p$=0.8), and changed globally over all occurrences to its noised version.
    \item The non-cognate \textbf{content} word noiser generates non-words using an HRL character ngram model as replacements for HRL content words. Note that this is the only noiser that models non-cognate processes. 
\end{itemize}

\section{Experimental details}

\subsection{Aya prompt}
\label{app:aya_prompt}

LLMs are known to be sensitive to the prompt used for the task \citep{anagnostidis2024susceptiblellmsinfluenceprompts}.

We evaluated \aya{} \ttt{off-the-shelf} on the following prompts, varying in their specification of the source language:
\begin{enumerate}
    \item \ttt{Translate into English:}
    
    \ttt{<source\_text>}
    \item \ttt{Translate from a dialect of \ttt{<hrl>} into English:}
    
    \ttt{<source\_text>}
    \item \ttt{Translate from \ttt{<flores\_code\_of\_lrl>} into English:}
    
    \ttt{<source\_text>}
\end{enumerate}

The first two prompts are chosen in accordance with our goal of translating potentially unknown dialects of the HRL. While prompts 2 and 3 generally give improvements on the baseline for known target dialects ($+1.3$, $+2.5$, $+1.3$, $-0.5$ mean BLEU improvements for Arabic, Romance, Turkic and Indic respectively) the best performing prompt differs by language. For the purpose of our study, using prompts 2 and 3 is non-ideal, since it introduces the confound of degradation stemming from a certain target being referred to as a dialect of another language, or model familiarity with the particular dialect name or code. This is relevant because our evaluation languages range over degrees of resourcedness and presence in \aya{}, as well as relatedness from the corresponding HRL. Further, note that also use our chosen prompt during instruction finetuning over noisy text for consistency, and it is unclear what language name or code we should use for the noised text for best transfer to all target dialects in this setting. In order to avoid the above confounds and for simplicity, we use the first prompt for all approaches and languages, both during training and evaluation.

\subsection{\mm{} tokenizer} 

\mm{} requires specification of source and target language tokens; the source language token is appended to the end of the input text during tokenization, and the target language token is provided to the decoder as its first token to specify the output language. We use CRL language tokens when they are supported by \mm{}, and the HRL language token otherwise. Note that the tokenizer itself is shared across all languages, as well as encoder parameters. While decoder parameters have language-family-specific components depending on the target language, this is irrelevant for us as we always have English as our target language.

\section{Bilingual Lexicon Collection for \dtm{}} 
\label{app:survey_of_lexicons}
Our \dtm{} approach uses CRL-HRL bilingual lexicons for an inference-time intervention. Ideally, we would like the lexicons to cover fully inflected words and all parts-of-speech. 

\subsection{Survey of lexicons, and challenges in collection}
See a list of the language-pair specific bilingual lexicons we considered in \autoref{tab:bilingual_lexicon_survey}: as shown here, many lexicons list lemmas rather than inflected words, rendering them unusable for our purpose. This is naturally particularly a problem for morphologically rich languages.

\paragraph{APIs and web apps} Many available sources for bilingual lexicons are APIs and web apps. These include but are not limited to Google Translate \footnote{\url{https://translate.google.com/}}, Microsoft Bing \footnote{\url{https://www.bing.com/translator}}, ModernMT API \footnote{\url{https://www.modernmt.com/translate}}, Alibaba Translate \footnote{\url{https://translate.alibaba.com/}}, tradukka \footnote{\url{https://tradukka.com/}}, freelang \footnote{\url{https://www.freelang.net/}}, From-to.io \footnote{\url{https://from-to.io/}}, iTranslate \footnote{\url{https://itranslate.com/}}, and Glosbe \footnote{\url{https://glosbe.com/}}. However:

\begin{itemize}
    \item \textbf{Querying is often problematic.} APIs are often blocked by some paywall and mining the databases of web apps may be considered unethical or illegal. This is the case for The Living Arabic Project \footnote{\url{https://www.livingarabic.com/}}, a database of endangered Arabic dialects.
    \item \textbf{The quality of such resources can be poor.} A study from the Kamusi Project found that Google Translate performed poorly on some of the study's languages, namely Plateau Malagasy, Ilocano, Samoan, and Maori \cite{kamusi_project}. Glosbe is a crowdsourced database, meaning the translations may not be accurate.
\end{itemize}

\paragraph{Programmatic readability} Some lexicons that exist as a PDF may have text that is difficult to extract. For example, the Peace Corps Tunisian Arabic Dictionary \citep{peacecorps_tunisian} is grainy, making it difficult to perform OCR.

\paragraph{English-centric resources}
While we could find many lexicons that translated from a low/mid-resource language into English, we had trouble finding lexicons that translated from a low/mid-resource language into a related non-English resource language (see Table \ref{tab:bilingual_lexicon_survey}). This underscores the existing bias toward English-centric resources and technologies in natural language processing \citep{rigouts-terryn-de-lhoneux-2024-exploratory}.

\subsection{Lexicon sizes}
\label{app:lexicon_sizes}
See Tables \ref{tab:bilingual_word_count_1}, \ref{tab:bilingual_word_count_2}, and \ref{tab:bilingual_word_count_3} for lexicon sizes of the lexicons we aggregated and used for each CRL-HRL pair, with a breakdown by function and content word counts. In these tables, we also report the word type coverage, a measure of how extensively the lexicon documents a LRL. Coverage is calculated as the percentage of unique words in the FloRes dev set/JHU Bible test set that are documented in the associated LRL lexicon.

Note that we only use publicly available data resources.

\section{Variant approaches and hyperparameter exploration}
\label{app:exp_variants}

\subsection{Choosing noising dials}
\label{app:choosing_noising_parameters}
The search space for $\theta^{p,m,f,c}$, i.e., the noising dials used to generate artificial dialectal data, is large. We conduct some small-scale experiments to understand the impact of this choice on the \mtd{} approaches.

\paragraph{Default choices for $\theta^{p,m,f,c}$}
We use the noiser framework as set up in \citet{bafna2024evaluating}. 
Within the Bayesian generative framework of these noisers, it is possible to compute the posteriors (MLE estimates) of $\theta^n$ independently for each noiser, given a real CRL-HRL pair.
\citet{bafna2024evaluating} compute and provide these posteriors over a number of CRL-HRL pairs for a few different language families, that include languages from four of the six language families that we work with.
Our default choices for these parameters depend on the observed range of these $\theta$-posteriors over 18 real language pairs.
For \mtd\ttt{-shell}, we take a simple average of these posteriors to give us a single $\theta^{p,m,f,c}$-radius, which serves as a reasonable default distance between a random CRL and HRL.
For \mtd\ttt{-cloud}, given that we generate artificial languages on several hyperspheres up to a maximum $\theta^{p,m,f,c}$-radius, we take maximums over the observed posteriors instead in each dimension, discarding some clear outliers.
While this yields reasonable defaults for the phonological, morphological, and function word noisers (which all model cognate processes), we find that this is not ideal for the content word noiser, as discussed next.

\paragraph{Setting $\theta^c$} Our intitial experiments indicate that setting $\theta^c$, i.e. the content word noiser dial in accordance with the procedure for calculating other noiser dials is suboptimal, and much lower $\theta^c$'s are better. This makes sense: the content word noiser is the only noiser that models \textit{non-cognate} processes, and introduces non-words with no connection to the source word into the noised text. This is naturally not helpful, since the introduced non-words have no systematic relationships with the target dialects, and there is no way for the model to generalize what it observes. Therefore, we set $\theta^c=0.001$, i.e., very low, and largely rely on and discuss the other three noisers for this work (as in \autoref{sec:discussion}).

\paragraph{Effect of hyperparameters on \mtd\ttt{-cloud}} We tried more aggressive noising for Indic, Turkic, and Haitian, using $\theta^{p,m,f,c}_{0.2,0.8,0.8,0.001}$ as the max-radius for \mtd\ttt{-cloud}; i.e. significantly increasing $\theta^p$ and $\theta^m$ (the $\theta^f$ default is already high at $0.8$).
This only gave minor variations on existing results for all language families and models. 
This is not very surprising: since \mtd{}\ttt{-cloud} samples from many radii through the hypersphere up to the max-radius, it is relatively less sensitive to the choice of the max-radius $\theta^{p,m,f,c}$.

\paragraph{Effect of hyperparameters on \mtd\ttt{-shell}} On the other hand, \mtd{}\ttt{-shell} only samples from the shell defined by the radius, and therefore might be more sensitive to this choice. We observe that it performs slightly better or worse than \mtd\ttt{-cloud} (see \autoref{mtd_m2mL_comet} and \autoref{mtd_aya-23-8b_comet}) depending on the language and model. Given a specific target language family, it may make sense to conduct a hyperparameter search for $\theta$ for optimal \mtd{}\ttt{-shell} performance. 
Our small-scale experiments with Indic with a few different $\theta$'s indicate only minor improvements from this search over our default parameters; however, this may vary by language family and model.
Given the intractability of a systematic hyperparameter search over 3-dimensional $\theta$-radius (excluding the non-cognate lexical noiser, i.e. $\theta^{p,m,f}$), we instead show that it is possible to approximate all cognate processes solely with $\theta^p$; we discuss this in \autoref{sec:discussion}.

\paragraph{Number of radii $N$ for \mtd{}\ttt{-cloud}} For the approach \mtd{}\ttt{-cloud}, we use $N$ radii at uniform intervals from $0$ to max-radius $\theta$; thus, $N$ defines the density of the cloud given a max-radius. Of course, the total amount of data used remains constant regardless of $N$ and $\theta$. We compute results for Indic for the choice of number of radii $K \in \{5, 10, 20\}$ and observe very minor differences. We use $N=10$ throughout the paper. Note that since $\theta$ is a probability, $N=10$ is already enough to approximate an effectively continuous cloud up to the max-radius. While there is no cost to increasing $N$, we observe no gains from doing so beyond a certain point.

\paragraph{Parametrizing \ttt{randaug}\ttt{-shell} and \ttt{-cloud}}

We can consider \ttt{randaug} to consist of a character-level and a word-level noiser.
The difference from our noisers lies only in the target generation: while our noisers choose linguistically plausible targets as described in \autoref{sec:method}, the \ttt{randaug} noisers choose random equivalents in accordance with previous literature.
The character-level noiser is analogous to the \mtd{} phonological noiser which affects characters, and the word-level noiser with the content word noiser which affects lexemes. 
For a fair comparison and in order to test the importance of linguistically-motivated target generation, we therefore set the dials for these noisers to be the same as the default dials for their analogous noisers, both for \ttt{-shell} and \ttt{-cloud} variants.
We use $\theta_{rc,rw}^{0.05,0.001}$ for \ttt{randaug-shell} and $\theta_{rc,rw}^{0.07,0.001}$ as the max-radius for \ttt{randaug-cloud}. 
For the latter, we use $K=10$ hyperspheres, consistent with \mtd\ttt{-cloud}.
Small tuning experiments over the Indic family yield minor variations in the baseline results obtained with these parameters.
Note that the morphological and function word noisers do not have random baseline equivalents in the literature: these are inherently linguistically-motivated and inspired by patterns of dialectal variation.

\subsection{Multiple HRLs} A language family may have more than one HRL, and the LRLs in the family may be closer to any of them. We experiment with using two HRLs as sources for noising, additionally using French for the Romance family, Uzbek for Turkic, and Tunisian for the Arabic family, along with the primary HRLs. We still maintain the total amount of finetuning data, but split it uniformly between our source HRLs. We choose Uzbek since it differs from Turkish in having a high amount of Russian-influenced and Persian-influenced vocabulary, similar to \ttt{crh\_Latn}. Similarly, we choose Tunisian as a representative of the Maghrebi Arabic dialects, hoping to help with Moroccan (\ttt{ary\_Arab}) which also belongs to this subfamily.
See \autoref{multiple_hrls} for the results for \aya{}. We see that this helps for some Romance dialects that share high similarities with French and Spanish (such as \ttt{oci\_Latn} and \ttt{glg\_Latn}). Similarly, for Arabic, this gives a considerable performance boost for Moroccan Arabic. However, we don't observe any improvements for the Turkic family except Uzbek itself. %

\subsection{Different datasets}
We experiment with different choices of base dataset for noising for Indic: namely, \ttt{indiccorp} \citep{doddapaneni-etal-2023-towards} and the NLLB dataset \citep{nllb2022}. We observed that the choice of dataset matters for the absolute performance of the finetuned model (\ttt{fthrl}): specifically, finetuning with \ttt{indiccorp} actually hurts the baseline performance of both models. However, the gains from the noising method over this finetuned baseline are consistent across these three datasets, for the Indic family. We present results on \ttt{indiccorp} in \autoref{indicorp_noising}); those on the NLLB dataset show similar trends (see \autoref{nllb_noising}).

\paragraph{Training hyperparameters} We observe that increasing either the amount of data used or number of epochs worsens ordinary HRL fine-tuning results slightly for CRLs, while \mtd{} approaches show small improvements for some languages. We fix these hyperparameters for all experiments (i.e. $100K$ sentences of bitext, a single epoch) for ease of comparison.

\subsection{Using CRL-Eng swapping} 
\label{app:crl_eng_swapping}
We experiment with switching CRL words directly into English instead of the HRL: as expected, this degrades performance considerably since the model now receives unnatural CRL-English code-mixed input. See \autoref{dtm_m2mL_eng} and \autoref{dtm_aya-23-8b_eng} for results.

\subsection{Enhancing noisers to model lexical variation: Semantic noise}
We would like to model different patterns of lexical usage among closely-related languages. For example, while the word ``book'' is ``pustak'' in Marathi and ``kitab'' in Hindi by common usage, ``pustak'' is also an entirely comprehensible Hindi word. We would like our noisers therefore to expose the model during finetuning to various lexical variants of a given concept as they might be realized in different dialects of a language family. We use WordNets in all our HRLs: IndoWordNet for Hindi \citep{bhattacharyya-2010-indowordnet}, TurkishWordNet \citep{bakay-etal-2021-turkish}, Arabic WordNet (AWN v2) \citep{regragui-etal-2016-arabic}, WordNet Bahasa \citep{noor-etal-2011-creating}, and Italian Wordnet \citep{roventini-etal-2000-italwordnet} to ``noise'' a given word with some probability $\theta^s$ to its (randomly chosen) synonym. Note that since these wordnets contain lemmas, whereas we typically encounter inflected forms in text, we first lemmatize the word to be noised, perform semantic noising, and then re-append its original inflections, to roughly maintain grammaticality. This process may introduce incidental noise at inflection boundaries, but is tolerable given our general goal of perturbing the data.

This noised and inflected synonym then undergoes the other noising processes as per usual, modeling the intuition that the diverging lexeme would still bear the effects of general phonological and morphological change in a given dialect. 

We set $\theta^s \in \{ 0.1, 0,3\}$, keeping other noising parameters constant; however, we do not see that this helps much. See language family means for \mm{} in \autoref{tab:using_different_thetas}.

\section{Function word identification for \dtm{}}
\label{app:func_word_collection}
\paragraph{HRL} \mtd{} requires function word identification in the HRL, for the function word lexical noiser. 
We use the same procedure for function word identification in the HRL as used by the lexical noiser in differential between functional and content words, consistent with \citet{bafna2024evaluating}, summarized in \autoref{app:noising_details}, to create a list of HRL function words.

\paragraph{CRLs} The \dtm{}\ttt{-func} approach only affects function words in the CRL input, and therefore requires function word identification for CRL text. Given that we have a set of HRL function words and HRL-CRL bilingual lexicons, we can identify function words in the CRL using projection of POS tags from the HRL.

\paragraph{Comparing against natural distribution}
The impact of our best-performing variant \dtm{}\ttt{-func} depends on the identification of function words in the CRL input as described above. We would like to evaluate the coverage of this method; however, we lack annotated data for the CRLs for this evaluation. Instead, assuming that language families share general distributive properties of function words, we compare the percentage of identified function words in our CRL input against the natural percentage of function words in the HRL (as a representative of the language family) in the UD corpus. See \autoref{natural_vs_changed_distribution}.
We observe that \dtm{} affects an expected percentage of words for Romance, Turkic, and Indic, but there is over-identification for Austronesian and under-identification for Arabic, possibly due to noisier automatic alignments for these families.

See \autoref{changed_words_across_langs} for a summary of the number of (1) content, (2) functional, and (3) content and functional combined (represented as ``all" in the table) words switched out in each \dtm {} approach.

\section{More examples}
\label{app:more_examples}
\begin{table*}[!ht]
\centering
    \resizebox{\textwidth}{!}{\includegraphics{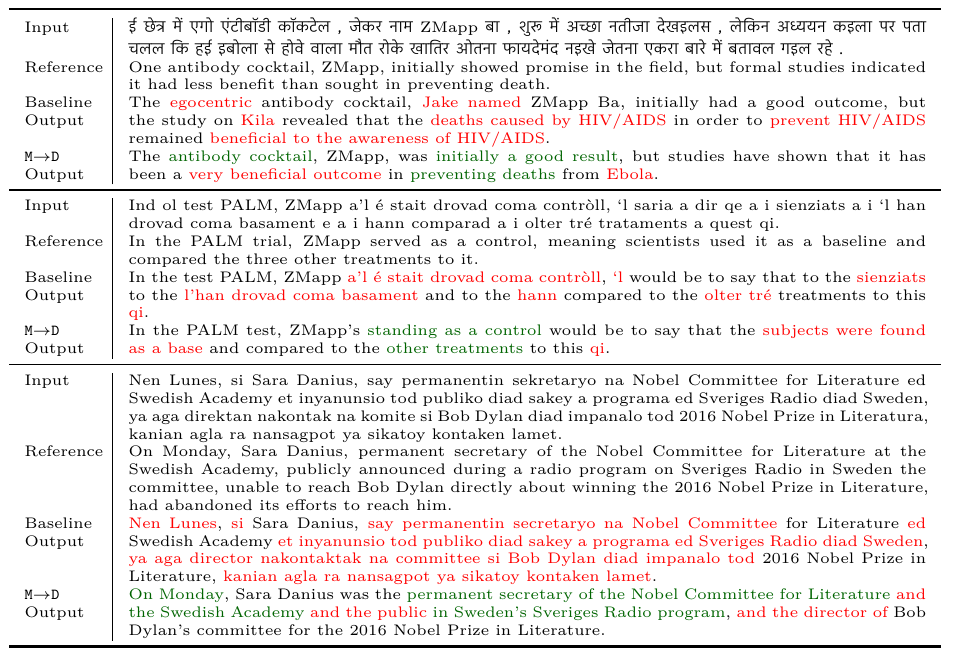}}
    \caption{Showing improvement of \mtd{} over \ttt{off-the-shelf}}
    \label{tab:mtd_examples}
\end{table*}

\begin{table*}[!ht]
\centering
    \resizebox{\textwidth}{!}{\includegraphics[scale=1]{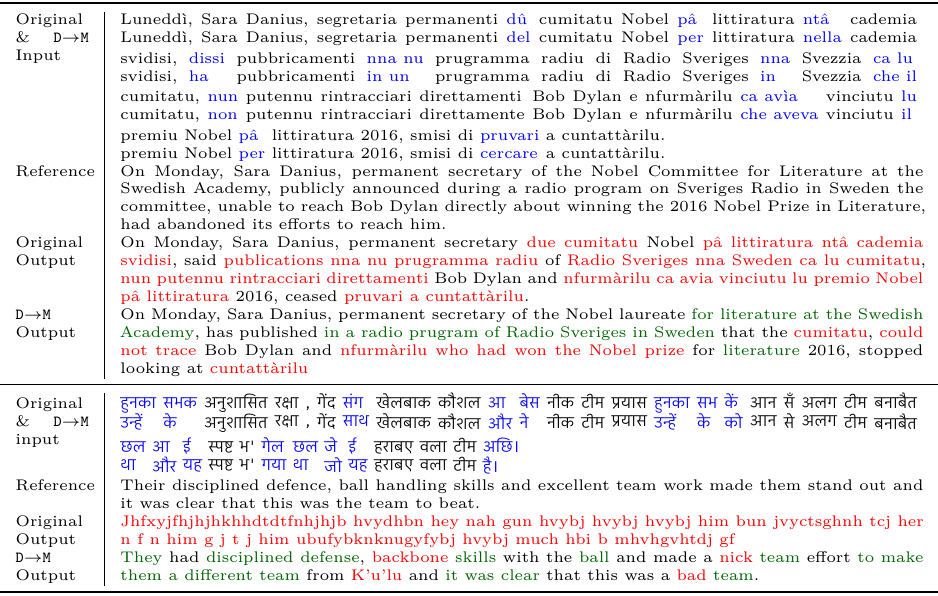}}
    \caption{Showing improvement of \dtm{} over \ttt{off-the-shelf}}
    \label{tab:dtm_examples}
\end{table*}

See examples of \mtd{} and \dtm{} outputs in \autoref{tab:mtd_examples} and \autoref{tab:dtm_examples}.

\newpage
\null
\newpage
\section{Wins and variance for approach paradigms}
\label{app:variance}
\begin{figure*}
    \centering
    \includegraphics[scale=0.7]{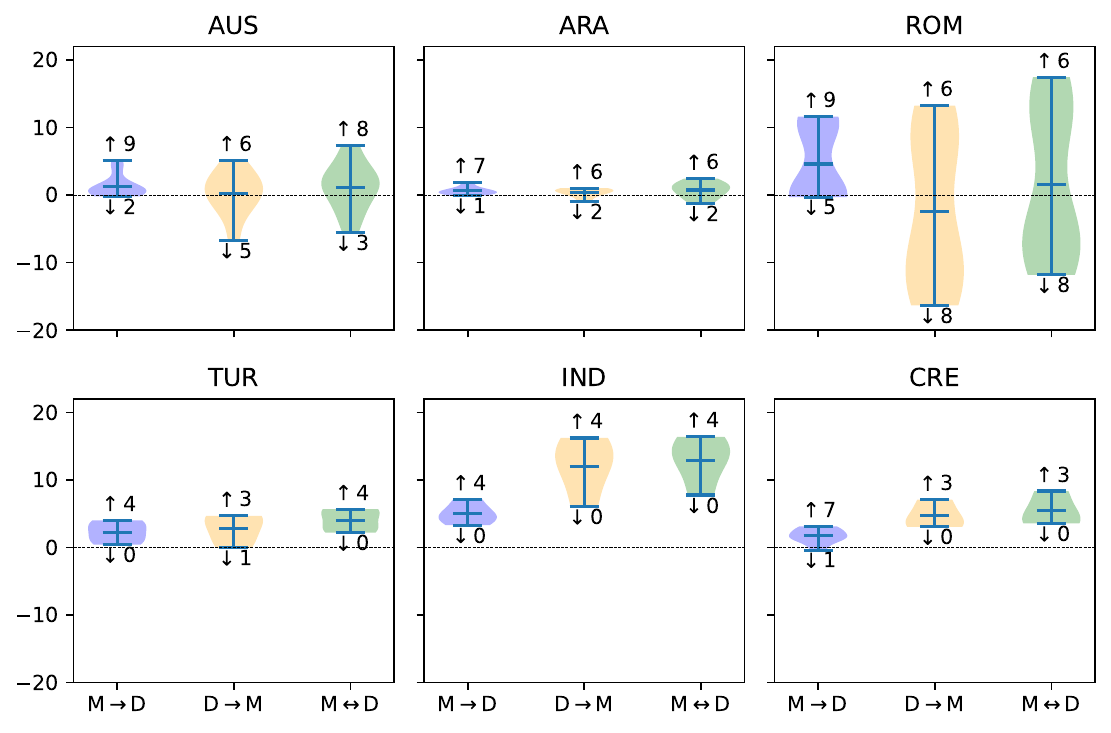}
    \caption{BLEU score improvements over the best baseline with \mm{} for all language families. $\uparrow$ and $\downarrow$: \# CRLs with positive/negative gains. \mtd{} gives more consistent positive gains. }
    \label{fig:variance_m2mL}
\end{figure*}

\begin{figure*}
    \centering
    \includegraphics[scale=0.7]{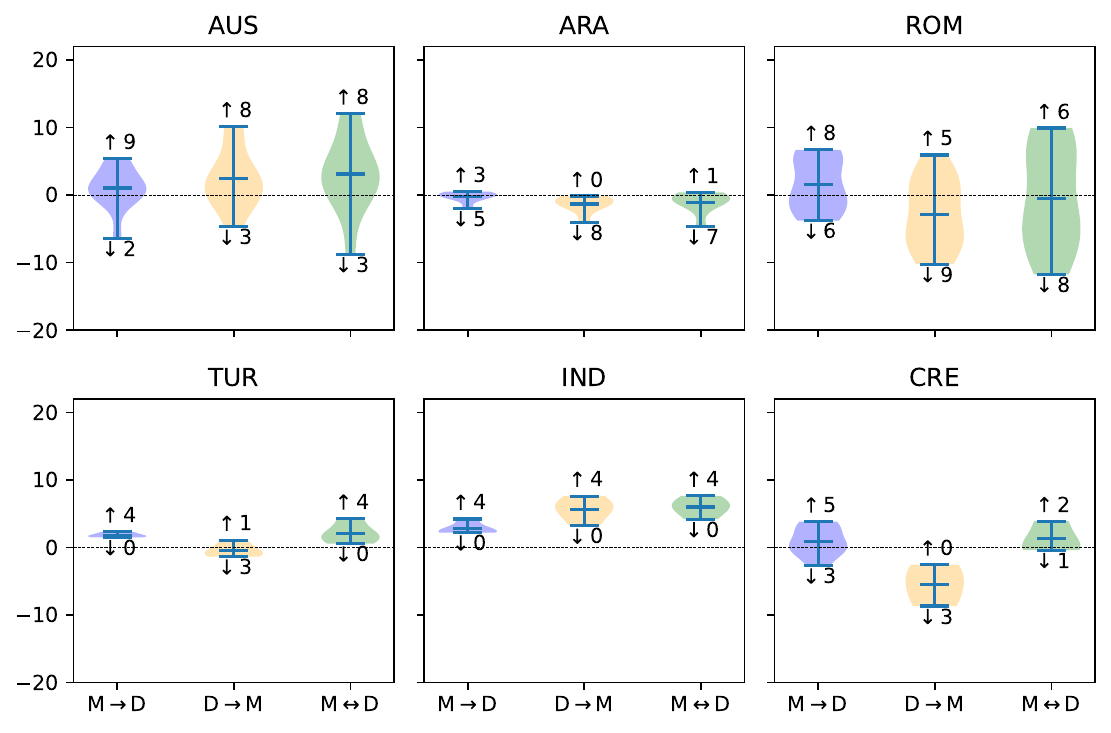}
    \caption{BLEU score improvements over the best baseline with \aya{} for all language families. $\uparrow$ and $\downarrow$: \# CRLs with positive/negative gains. \mtd{} gives more consistent positive gains. }
    \label{fig:variance_aya}
\end{figure*}

See \autoref{fig:variance_m2mL} and \autoref{fig:variance_aya} for a comparison of win rates over baseline and improvement distributions for different paradigms for \mm{} and \aya{} respectively, for all language families.

\section{Manual Evaluation of Function Word Lexicons}
\label{app:manual_evaluation}

\begin{table}[hbtp]
\centering
\small
\resizebox{0.45\textwidth}{!}{
\begin{tabular}{lccc}
\toprule
\textbf{lang} & \textbf{FW acc. (\%)} & \textbf{FW Id. (\%)} & \textbf{Gen. acc. (\%)} \\
\midrule
fra\_ita & 67.6 & 67.0 & 60.4 \\
bho\_hin & 97.9 & 51.6 & 74.7 \\
arz\_arb & 84.0 & 32.9 & 38.2 \\
azj\_tur & 84.9 & 74.5 & 65.3 \\
Mean    & 83.6 & 56.5 & 59.7 \\
\bottomrule
\end{tabular}}
\caption{Function word (FW) accuracy (\%), function word identification accuracy (FW Id, \%), general translation accuracy (\%) in function word lexicons}
\label{tab:manual_eval_lexicons}
\end{table}

The evaluation was conducted by three of our authors, fluent in Turkish (tur), Hindi (hin), Arabic (arb), Egyptian (ary), and French (fra), with a working understanding of the respective CRLs and Italian (ita). See \autoref{tab:manual_eval_lexicons}.

\section{Training and evaluation details}
\label{app:train_eval_details}

\subsection{Compute}
Our noising approaches and baselines require finetuning: we had 2 models x 8 approaches x 6 HRLs = 96 finetuning runs. Each experiment required about 3 hours on a single A100, totaling 288 GPU hours. 

We conducted evaluation for 54 languages (49 CRLs + 6 HRLs) x 15 total approaches = 810 evaluation runs. Each evaluation took 30 minutes on a single GTX 1080 machine, totaling 405 GPU hours.

These calculations do not include development and initial experiments, and ablation studies.

\subsection{Evaluation}
We tested our models on the \ttt{devtest} splits of the FloRes-200 dataset, using BLEU (HuggingFace \ttt{evaluate} wrapper: mixed case, tokenization with \ttt{sacrebleu} \ttt{13a}) as well as COMET (\ttt{Unbabel/wmt22-comet-da}).

\section{Results by language and approach}
\label{app:more_results}

We summarize our results in the following.

\begin{itemize}
    \item \autoref{bleu_langfam} and \autoref{COMET_langfam} detail the average BLEU and COMET scores respectively for each approach within the \mtd, \dtm, and \combined{} paradigms for each language family. \autoref{comet_langfam_low_baseline} provides COMET score means for low-baseline CRLs (<25 BLEU) per language family.
    \item \autoref{m2mL_bleu_best}, \autoref{aya-23-8b_bleu_best}, \autoref{m2mL_COMET_best}, and \autoref{aya-23-8b_COMET_best} present the  BLEU and COMET scores of best-performing variant each of the baseline, \mtd, \dtm, and \combined{} approaches, for \mm{} and \aya{}. Note that the best approach is chosen on a family-wide basis, in order to recommend a general strategy for a given language family, by comparing means over the language family for each variant, and may not reflect the best scores for an individual language in that paradigm (similarly for the best baseline). \textbf{This is different from how best variants are chosen for \autoref{fig:bar_charts}, \autoref{fig:results_variance}, \autoref{fig:variance_m2mL}, and \autoref{fig:variance_aya}}, which consider best performing paradigms variants on a per-language basis. 
\end{itemize}

The following tables provide a more detailed view of the performance of each paradigm.

\begin{itemize}
    \item \textbf{\mtd}: \autoref{mtd_m2mL_bleu}, \autoref{mtd_aya-23-8b_bleu},  \autoref{mtd_m2mL_comet}. and \autoref{mtd_aya-23-8b_comet} detail the BLEU and COMET performance for each language in our language family families using each approach in the \mtd {} paradigm. \autoref{tab:mtd_examples} provides a qualitative example of how translations improved following the \mtd {} paradigm.
    \item \textbf{\dtm}: \autoref{dtm_m2mL_bleu}, \autoref{dtm_aya-23-8b_bleu}, \autoref{dtm_m2mL_comet}, and \autoref{dtm_aya-23-8b_comet} detail the BLEU and COMET performance for each language in our language family families using each approach in the \dtm {} paradigm. \autoref{tab:dtm_examples} provides a qualitative example of how translations improved following the \dtm {} paradigm.
    \item \textbf{\combined}: \autoref{mtd+dtm_m2mL_bleu}, \autoref{mtd+dtm_aya-23-8b_bleu}, \autoref{mtd+dtm_m2mL_comet}, and \autoref{mtd+dtm_aya-23-8b_comet} detail the BLEU and COMET performance for each language in our language family families using each approach in the \combined {} paradigm.
\end{itemize}

\section{Use of AI assistants}
We used GitHub copilot for coding assistance. No AI assistants were used for any writing purposes.

\begin{table*}[htbp]
	\centering
	\resizebox{1\textwidth}{!}{
\begin{tabular}{l l | r r r r r r | r | r r r r r r | r}
		\toprule
		& & \multicolumn{7}{c|}{\textbf{\mm}} & \multicolumn{7}{c}{\textbf{\aya}} \\
		\midrule
		   & & AUS (9) & ARA (7) & ROM (6) & TUR (4) & IND (4) & CRE (8) & Mean & AUS (9) & ARA (1) & ROM (5) & TUR (4) & IND (4) & CRE (9) & Mean  \\
\midrule
\textbf{Baselines} & \ttt{\textbf{off-the-shelf}} & 14.9 & 21.2 & 26.6 & 5.0 & 12.6 & 5.9 & 14.3 & \underline{11.9} & \textbf{\underline{31.5}} & 29.5 & 7.9 & \underline{20.4} & 7.3 & \underline{18.1}  \\
& \ttt{\textbf{fthrln}} & \underline{+0.3} & \underline{+0.3} & \underline{+0.4} & -0.1 & \underline{+0.4} & \underline{+0.9} & \underline{+0.4} & -1.0 & -0.7 & -0.1 & \underline{+1.2} & -1.8 & \underline{+2.3} & -0.0  \\
& \ttt{\textbf{ftrandaug}} & +0.2 & +0.2 & +0.2 & \underline{+0.2} & \underline{+0.4} & \underline{+0.9} & +0.3 & -0.9 & -0.5 & \underline{+0.2} & +1.1 & -1.8 & +2.1 & \underline{+0.0}  \\
\midrule
\mtd & \ttt{\textbf{-shell}} & \textbf{\underline{+1.7}} & \underline{+1.0} & \textbf{\underline{+5.1}} & \underline{+2.5} & +5.0 & \underline{+2.7} & \underline{+3.0} & +1.1 & -0.2 & +2.4 & +2.9 & \underline{+2.8} & +3.0 & +2.0  \\
& \ttt{\textbf{-cloud}} & +1.1 & +0.7 & +3.9 & +1.6 & \underline{+5.4} & \underline{+2.7} & +2.6 & \underline{+1.2} & \underline{-0.0} & \textbf{\underline{+2.5}} & \textbf{\underline{+3.2}} & +2.7 & \textbf{\underline{+3.5}} & \textbf{\underline{+2.2}}  \\
\midrule
\dtm & \ttt{\textbf{-cont}} & \underline{-1.5} & -2.0 & -7.1 & +1.4 & +0.3 & +2.5 & -1.1 & -0.1 & -4.3 & -7.6 & -0.6 & -1.1 & -3.8 & -2.9  \\
\bottomrule
	 \end{tabular}}
	\caption{Comparison of language family BLEU means by model. Performance gains/loss are relative to \ttt{off-the-shelf}. The overall best score is \textbf{bolded} and the best score in each paradigm is \underline{underlined}.}
	\label{bleu_langfam}
\end{table*}

\begin{table*}[htbp]
	\centering
	\resizebox{1\textwidth}{!}{
		\begin{tabular}{l l | r r r r r r | r | r r r r r r | r}
			 \toprule
             & & \multicolumn{7}{c|}{\textbf{\mm}} & \multicolumn{7}{c}{\textbf{\aya}} \\
            \midrule

               & & AUS (9) & ARA (7) & ROM (6) & TUR (4) & IND (4) & CRE (8) & Mean & AUS (9) & ARB (1) & ROM (5) & TUR (4) & IND (4) & CRE (9) & Mean  \\
			 \midrule
Baselines & \ttt{\textbf{off-the-shelf}} & 52.9 & \underline{72.6} & \underline{67.9} & \underline{47.6} & \underline{64.2} & \underline{47.6} & \underline{58.8} & \underline{55.2} & 81.7 & 74.0 & 60.7 & \underline{76.5} & 47.6 & 66.0 \\
 & \ttt{\textbf{fthrl}} & \underline{+0.7} & -0.1 & -0.6 & -0.6 & \underline{+0.0} & -0.3 & -0.1 & -2.0 & +0.2 & +0.9 & \underline{+4.3} & -1.7 & \underline{+6.2} & +1.3 \\
 & \ttt{\textbf{randaug-shell}} & +0.1 & -0.5 & -0.2 & -0.3 & -0.0 & -0.3 & -0.2 & -1.6 & \underline{+0.3} & \underline{+1.5} & +4.3 & -1.8 & +5.7 & \underline{+1.4} \\
\midrule
\mtd & \ttt{\textbf{-shell}} & \underline{+4.3} & \textbf{\underline{+1.3}} & \textbf{\underline{+8.3}} & \underline{+8.0} & +8.0 & \textbf{\underline{+3.6}} & \underline{+5.6} & +3.8 & +1.0 & +5.1 & \textbf{\underline{+8.3}} & +2.8 & +7.5 & +4.8 \\
 & \ttt{\textbf{-cloud}} & +2.9 & +0.8 & +7.0 & +4.4 & \underline{+8.7} & +3.3 & +4.5 & \underline{+4.1} & \textbf{\underline{+1.1}} & \textbf{\underline{+5.2}} & +8.2 & \underline{+3.0} & \textbf{\underline{+7.8}} & \underline{+4.9} \\
\midrule
\dtm & \ttt{\textbf{-cont}} & +0.3 & -2.7 & -8.7 & +5.2 & +0.8 & - & +1.2 & \underline{+0.5} & -4.1 & -8.6 & -3.7 & -1.3 & - & +0.2 \\
\bottomrule
		 \end{tabular}
	 }
	\caption{Comparison of language family COMET means by model. Performance gains/losses are relative to \ttt{off-the-shelf}. The overall best score is \textbf{bolded} and best score in each paradigm is \underline{underlined}.}
	\label{COMET_langfam}
\end{table*}

\begin{table*}[htbp]
	\centering
	\resizebox{1\textwidth}{!}{
		\begin{tabular}{l l | r r r r r r | r | r r r r r r | r}
			 \toprule
             & & \multicolumn{7}{c|}{\textbf{\mm}} & \multicolumn{7}{c}{\textbf{\aya}} \\
            \midrule

               & & AUS (9) & ARA (7) & ROM (6) & TUR (4) & IND (4) & CRE (8) & Mean & AUS (9) & ARB (1) & ROM (5) & TUR (4) & IND (4) & CRE (9) & Mean  \\
			 \midrule
Baselines & \ttt{\textbf{off-the-shelf}} & 46.9 & \underline{71.5} & \underline{44.7} & \underline{47.6} & 64.2 & \underline{47.6} & \underline{53.8} & \underline{49.5} & 74.0 & 58.4 & 60.7 & \underline{76.5} & 47.6 & 61.1 \\
 & \ttt{\textbf{fthrl}} & \underline{+0.7} & -0.1 & -1.3 & -0.6 & \underline{+0.0} & -0.3 & -0.2 & -2.0 & +0.2 & +2.8 & \underline{+4.3} & -1.7 & \underline{+6.2} & +1.6 \\
 & \ttt{\textbf{randaug-shell}} & +0.2 & -0.5 & -0.3 & -0.3 & -0.0 & -0.3 & -0.2 & -1.6 & \underline{+0.6} & \underline{+4.0} & +4.3 & -1.8 & +5.7 & \underline{+1.9} \\
\midrule
\mtd & \ttt{\textbf{-shell}} & \underline{+5.1} & \textbf{\underline{+1.5}} & \underline{+19.4} & \underline{+8.0} & +8.0 & \textbf{\underline{+3.6}} & \underline{+7.6} & +5.3 & +2.7 & +12.7 & \textbf{\underline{+8.3}} & +2.8 & +7.5 & +6.6 \\
 & \ttt{\textbf{-cloud}} & +3.4 & +1.0 & +16.5 & +4.4 & \underline{+8.7} & +3.3 & +6.2 & \underline{+5.5} & \textbf{\underline{+2.9}} & \underline{+12.7} & +8.2 & \underline{+3.0} & \textbf{\underline{+7.8}} & \underline{+6.7} \\
\midrule
\dtm & \ttt{\textbf{-cont}} & +2.0 & -3.1 & +2.8 & +5.2 & +0.8 & - & +2.8 & +1.9 & -6.5 & -2.1 & -3.7 & -1.3 & - & +0.4 \\
\bottomrule
		 \end{tabular}
	 }
	\caption{Comparison of language family COMET means by model where baseline < 25. Performance gains/loss are relative to \ttt{off-the-shelf}. The overall best score for each language is \textbf{bolded} and the best score in each paradigm is \underline{underlined}. Numbers in parenthesis represent the number of languages within the model and language family whose baseline \ttt{off-the-shelf} BLEU score was less than 25.
    }
	\label{comet_langfam_low_baseline}
\end{table*}

\begin{table*}[htbp]
	\tiny
	\centering
	\resizebox{0.5\textwidth}{!}{
		\begin{tabular}{l r r r r}
			 \toprule
			 \textbf{Language} & \textbf{Best Baseline} & \textbf{Best \mtd} & \textbf{Best \dtm} & \textbf{Best \combined} \\
			 \midrule
\midrule
\multicolumn{4}{l}{\textbf{Austronesian Language Family}} \\
\midrule
ind\_Latn (Indonesian) & \textbf{\underline{38.9}} & -0.5 & - & -  \\
jav\_Latn (Javanese) & 23.3 & \textbf{\underline{+0.7}} & -0.2 & \textbf{\underline{+0.7}}  \\
sun\_Latn (Sundanese) & 23.5 & +0.8 & +0.4 & \textbf{\underline{+0.9}}  \\
smo\_Latn (Samoan) & 2.4 & -0.2 & +2.0 & \textbf{\underline{+2.7}}  \\
mri\_Latn (Maori) & 2.5 & +0.4 & +2.1 & \textbf{\underline{+2.6}}  \\
ceb\_Latn (Cebuano) & 22.6 & \textbf{\underline{+1.1}} & -3.1 & -2.3  \\
\bottomrule
		 \end{tabular}
	 }
	\caption{\mm{} BLEU for best performing \mtd, \dtm, \combined{} by language. Performance gains/loss are relative to \ttt{Best Baseline}.}
	\label{m2mL_bleu_best}
\end{table*}

\begin{table*}[htbp]
	\tiny
	\centering
	\resizebox{0.5\textwidth}{!}{
		\begin{tabular}{l r r r r}
			 \toprule
			 \textbf{Language} & \textbf{Best Baseline} & \textbf{Best \mtd} & \textbf{Best \dtm} & \textbf{Best \combined} \\
			 \midrule
\midrule
\multicolumn{4}{l}{\textbf{Austronesian Language Family}} \\
\midrule
ind\_Latn (Indonesian) & \textbf{\underline{41.5}} & -1.0 & - & -  \\
jav\_Latn (Javanese) & 15.2 & +4.9 & +9.6 & \textbf{\underline{+11.2}}  \\
sun\_Latn (Sundanese) & 12.2 & +5.4 & +10.2 & \textbf{\underline{+12.1}}  \\
smo\_Latn (Samoan) & 3.2 & +1.1 & +1.1 & \textbf{\underline{+2.2}}  \\
mri\_Latn (Maori) & 4.8 & +0.1 & +0.4 & \textbf{\underline{+2.0}}  \\
ceb\_Latn (Cebuano) & \textbf{\underline{14.0}} & -0.6 & -0.0 & -0.3  \\
\bottomrule
		 \end{tabular}
	 }
	\caption{\aya{} BLEU for best performing \mtd, \dtm, \combined{} by language. Performance gains/loss are relative to \ttt{Best Baseline}.}
	\label{aya-23-8b_bleu_best}
\end{table*}

\begin{table}[htbp]
	\tiny
	\centering
	\resizebox{0.45\textwidth}{!}{
		\begin{tabular}{l r r r r}
			 \toprule
			 \textbf{Language} & \textbf{Best Baseline} & \textbf{Best \mtd} & \textbf{Best \dtm} & \textbf{Best \combined} \\
			 \midrule
\midrule
\multicolumn{5}{l}{\textbf{Austronesian Language Family}} \\
\midrule
ind\_Latn (Indonesian) & 87.3 & +0.0 & - & - \\
\midrule
jav\_Latn (Javanese) & 70.9 & +1.2 & -1.5 & -0.4 \\
sun\_Latn (Sundanese) & 72.3 & +1.3 & -3.4 & -2.4 \\
smo\_Latn (Samoan) & 27.6 & +2.8 & +13.3 & +12.2 \\
mri\_Latn (Maori) & 25.2 & +6.5 & +17.6 & +17.2 \\
ceb\_Latn (Cebuano) & 64.7 & +1.4 & -10.3 & -2.5 \\
\bottomrule
		 \end{tabular}
	 }
	\caption{m2mL's COMET for best performing \mtd, \dtm, \combined  {} by language. Performance gains/loss are relative to \ttt{Best Baseline}.}
	\label{m2mL_COMET_best}
\end{table}

\begin{table}[htbp]
	\tiny
	\centering
	\resizebox{0.45\textwidth}{!}{
		\begin{tabular}{l r r r r}
			 \toprule
			 \textbf{Language} & \textbf{Best Baseline} & \textbf{Best \mtd} & \textbf{Best \dtm} & \textbf{Best \combined} \\
			 \midrule
\midrule
\multicolumn{5}{l}{\textbf{Austronesian Language Family}} \\
\midrule
ind\_Latn (Indonesian) & 88.5 & -0.1 & - & - \\
\midrule
jav\_Latn (Javanese) & 67.5 & +6.2 & +1.1 & +6.6 \\
sun\_Latn (Sundanese) & 63.7 & +9.0 & +1.7 & +8.9 \\
smo\_Latn (Samoan) & 35.4 & +6.2 & +3.9 & +11.4 \\
mri\_Latn (Maori) & 40.5 & +2.2 & +3.6 & +9.0 \\
ceb\_Latn (Cebuano) & 61.1 & +0.3 & -2.2 & +0.3 \\
\bottomrule
		 \end{tabular}
	 }
	\caption{aya-23-8b's COMET for best performing \mtd, \dtm, \combined  {} by language. Performance gains/loss are relative to \ttt{Best Baseline}.}
	\label{aya-23-8b_COMET_best}
\end{table}

\begin{table*}[htbp]
	\tiny
	\centering
	\resizebox{0.6\textwidth}{!}{
		\begin{tabular}{l | r r r r | r r}
			 \toprule
			 \textbf{Language} & \ttt{off-the-shelf} & \ttt{fthrln} & \ttt{ftrandaug} & \ttt{ftrandaug-cloud} & \ttt{-shell} & \ttt{-cloud}  \\
			 \midrule
\midrule
\multicolumn{4}{l}{\textbf{Austronesian Language Family}} \\
\midrule
ind\_Latn (Indonesian) & \textbf{\underline{38.9}} & -0.1 & -0.4 & -0.4 & \underline{-0.5} & -0.7  \\
\midrule
jav\_Latn (Javanese) & 22.9 & \underline{+0.4} & +0.0 & -0.1 & \textbf{\underline{+1.1}} & +1.0  \\
sun\_Latn (Sundanese) & 22.9 & \underline{+0.6} & +0.1 & +0.1 & \textbf{\underline{+1.4}} & +0.8  \\
smo\_Latn (Samoan) & 2.3 & \textbf{\underline{+0.1}} & \textbf{\underline{+0.1}} & \textbf{\underline{+0.1}} & -0.3 & \underline{-0.1}  \\
mri\_Latn (Maori) & 2.3 & \underline{+0.2} & \underline{+0.2} & \underline{+0.2} & \textbf{\underline{+0.6}} & +0.3  \\
ceb\_Latn (Cebuano) & 22.2 & \underline{+0.4} & +0.2 & +0.3 & \textbf{\underline{+1.5}} & +0.8  \\
\bottomrule
		 \end{tabular}
	 }
	\caption{\mtd{} BLEU scores by language for the model \mm. Performance gains/losses are relative to \ttt{off-the-shelf}. Averages for each \mtd{} approach are computed for each language family as well as for the general body of languages studied. These averages are recomputed for languages whose \ttt{off-the-shelf} BLEU score < 25. The overall best score for each language is \textbf{bolded} and the best score in each paradigm is \underline{underlined}.}
	\label{mtd_m2mL_bleu}
\end{table*}

\begin{table*}[htbp]
	\tiny
	\centering
	\resizebox{0.6\textwidth}{!}{
		\begin{tabular}{l | r r r r | r r}
			 \toprule
			 \textbf{Language} & \ttt{off-the-shelf} & \ttt{fthrln} & \ttt{ftrandaug} & \ttt{ftrandaug-cloud} & \ttt{-shell} & \ttt{-cloud}  \\
			 \midrule
\midrule
\multicolumn{4}{l}{\textbf{Austronesian Language Family}} \\
\midrule
ind\_Latn (Indonesian) & \textbf{\underline{41.5}} & -0.4 & -0.3 & -0.4 & \underline{-1.0} & -1.1  \\
\midrule
jav\_Latn (Javanese) & 14.2 & \underline{+1.0} & +0.4 & +0.8 & \textbf{\underline{+5.9}} & +5.5  \\
sun\_Latn (Sundanese) & 11.0 & +0.8 & \underline{+1.2} & +0.9 & \textbf{\underline{+6.6}} & \textbf{\underline{+6.6}}  \\
smo\_Latn (Samoan) & \underline{3.2} & -0.6 & -0.4 & -0.5 & \textbf{\underline{+1.1}} & \textbf{\underline{+1.1}}  \\
mri\_Latn (Maori) & \underline{4.8} & -1.5 & -1.2 & -1.4 & +0.0 & \textbf{\underline{+0.1}}  \\
ceb\_Latn (Cebuano) & \textbf{\underline{14.0}} & -2.3 & -2.1 & -2.2 & -1.1 & \underline{-0.6}  \\
\bottomrule
		 \end{tabular}
	 }
	\caption{\mtd{} BLEU scores by language for the model \aya. Performance gains/losses are relative to \ttt{off-the-shelf}. Averages for each \mtd{} approach are computed for each language family as well as for the general body of languages studied. These averages are recomputed for languages whose \ttt{off-the-shelf} BLEU score < 25. The overall best score for each language is \textbf{bolded} and the best score in each paradigm is \underline{underlined}.}
	\label{mtd_aya-23-8b_bleu}
\end{table*}

\begin{table*}[htbp]
	\tiny
	\centering
	\resizebox{0.6\textwidth}{!}{
		\begin{tabular}{l | r r r r | r r}
			 \toprule
			 \textbf{Language} & \ttt{off-the-shelf} & \ttt{fthrl} & \ttt{randaug-shell} & \ttt{randaug-cloud} & \ttt{-shell} & \ttt{-cloud}  \\
			 \midrule
\midrule
\multicolumn{4}{l}{\textbf{Austronesian Language Family}} \\
\midrule
ind\_Latn (Indonesian) & 87.4 & -0.1 & -0.2 & -0.2 & -0.1 & -0.1 \\
\midrule
jav\_Latn (Javanese) & 70.1 & +0.9 & -0.0 & -0.1 & +2.1 & +1.6 \\
sun\_Latn (Sundanese) & 71.6 & +0.7 & -0.2 & -0.3 & +2.0 & +1.6 \\
smo\_Latn (Samoan) & 27.4 & +0.2 & +0.0 & +0.0 & +3.1 & +1.9 \\
mri\_Latn (Maori) & 25.7 & -0.6 & -0.6 & -0.6 & +5.9 & +3.2 \\
ceb\_Latn (Cebuano) & 64.2 & +0.5 & -0.4 & -0.4 & +1.8 & +1.1 \\
\bottomrule
		 \end{tabular}
	 }
	\caption{\mtd{} COMET scores by language for the model \mm. Performance gains/losses are relative to \ttt{off-the-shelf}. Averages for each \mtd{} approach are computed for each language family as well as for the general body of languages studied. These averages are recomputed for languages whose \ttt{off-the-shelf} BLEU score < 25.}
	\label{mtd_m2mL_comet}
\end{table*}

\begin{table*}[htbp]
	\tiny
	\centering
	\resizebox{0.7\textwidth}{!}{
		\begin{tabular}{l | r r r r | r r}
			 \toprule
			 \textbf{Language} & \ttt{off-the-shelf} & \ttt{fthrl} & \ttt{randaug-shell} & \ttt{randaug-cloud} & \ttt{-shell} & \ttt{-cloud}  \\
			 \midrule
\midrule
\multicolumn{4}{l}{\textbf{Austronesian Language Family}} \\
\midrule
ind\_Latn (Indonesian) & 88.5 & +0.1 & +0.1 & +0.0 & -0.0 & -0.1 \\
\midrule
jav\_Latn (Javanese) & 67.5 & +0.8 & +0.8 & +1.0 & +6.2 & +6.2 \\
sun\_Latn (Sundanese) & 63.7 & +0.9 & +1.3 & +1.2 & +8.7 & +9.0 \\
smo\_Latn (Samoan) & 35.4 & -2.6 & -2.0 & -2.5 & +6.3 & +6.2 \\
mri\_Latn (Maori) & 40.5 & -6.7 & -5.6 & -6.1 & +1.7 & +2.2 \\
ceb\_Latn (Cebuano) & 61.1 & -3.7 & -3.5 & -3.7 & +0.1 & +0.3 \\
\bottomrule
		 \end{tabular}
	 }
	\caption{\mtd{} COMET scores by language for the model \aya. Performance gains/losses are relative to \ttt{off-the-shelf}. Averages for each \mtd{} approach are computed for each language family as well as for the general body of languages studied. These averages are recomputed for languages whose \ttt{off-the-shelf} BLEU score < 25.}
	\label{mtd_aya-23-8b_comet}
\end{table*}

\begin{table}[!ht]
        \centering
	\resizebox{0.4\textwidth}{!}{
    \small
	\begin{tabular}{l r r}
		\toprule
            & \textbf{\ttt{-cloud}} & \textbf{\ttt{-cloud-multsrcs}} \\
            \midrule
            \multicolumn{3}{l}{\textbf{Arabic Language Family (FloRes Dataset)}} \\
            \midrule
            acm\_Arab & 30.4 & 30.5 \\
            acq\_Arab & 31.4 & 31.8 \\
            aeb\_Arab & 25.4 & 26.0 \\
            ajp\_Arab & 33.5 & 35.2 \\
            apc\_Arab & 30.3 & 32.4 \\
            arb\_Arab & 37.1 & 36.0 \\
            ars\_Arab & 36.1 & 35.2 \\
            ary\_Arab & 19.7 & 23.8 \\
            arz\_Arab & 27.9 & 29.4 \\
            \midrule
            \textbf{Average} & 30.2 & 31.1 \\
            \midrule
            \multicolumn{3}{l}{\textbf{Romance Language Family}} \\
            \midrule
            ast\_Latn & 31.6 & 33.4 \\
            cat\_Latn & 33.4 & 35.8 \\
            fra\_Latn & 40.9 & 41.4 \\
            fur\_Latn & 24.4 & 24.1 \\
            glg\_Latn & 34.1 & 36.4 \\
            ita\_Latn & 32.0 & 33.5 \\
            lij\_Latn & 25.6 & 25.0 \\
            lmo\_Latn & 26.2 & 25.5 \\
            oci\_Latn & 36.5 & 37.3 \\
            por\_Latn & 43.8 & 45.2 \\
            ron\_Latn & 38.2 & 40.4 \\
            scn\_Latn & 26.1 & 25.5 \\
            spa\_Latn & 30.7 & 30.9 \\
            srd\_Latn & 24.1 & 23.5 \\
            vec\_Latn & 32.7 & 31.0 \\
            \midrule
            \textbf{Average} & 32.0 & 32.6 \\
            \midrule
            \multicolumn{3}{l}{\textbf{Turkic Language Family}} \\
            \midrule
            azj\_Latn & 11.3 & 11.6 \\
            crh\_Latn & 17.0 & 16.4 \\
            tuk\_Latn & 9.4	& 9.7 \\
            tur\_Latn & 29.2 & 27.7 \\
            uzn\_Latn & 7.6	& 9.2 \\
            \midrule
            \textbf{Average} & 14.9 & 14.9 \\
            \bottomrule
	\end{tabular}
	}
	\caption{Summary of adopting multiple HRLs as sources for the \mtd{} approach on the model \aya. These results are computed over a 300 sample-subset of the FloRes-200 test set, for \aya{}}
	\label{multiple_hrls}
\end{table}

\begin{table}[!ht]
    \centering
    \resizebox{0.45\textwidth}{!}{
    \begin{tabular}{l r r r}
        \toprule
        \textbf{Language} & \textbf{\ttt{off=the-shelf}} & \textbf{\ttt{fthrl}} & \textbf{\ttt{-shell}} \\
        \midrule
        bho\_Deva & 10.1 & 10.0 & 13.3 \\
        hin\_Deva & 34.8 & 33.4 & 33.3 \\
        hne\_Deva & 14.5 & 13.2 & 19.5 \\
        mag\_Deva & 17.4 & 16.0 & 21.3 \\
        mai\_Deva & 9.6	& 7.1 & 16.1 \\
        \midrule
        \textbf{Average} & 17.3 & 15.9 & 20.7 \\
        \bottomrule
    \end{tabular}}
    \caption{Using \texttt{indicorp} to noise for languages in the Indic family. The \mm{} model was tested. These results are computed over a 300 sample-subset of the FloRes-200 test set.}
    \label{indicorp_noising}
\end{table}

\begin{table}[!ht]
    \resizebox{0.45\textwidth}{!}{
    \begin{tabular}{l r r r}
        \toprule
            \textbf{Language} & \textbf{\ttt{off-the-shelf}} & \textbf{\ttt{fthrl}} & \textbf{\ttt{-shell}} \\
        \midrule
        bho\_Deva & 10.1 & 10.0 & 13.3 \\
        hin\_Deva & 34.8 & 33.4 & 33.3 \\
        hne\_Deva & 14.5 & 13.2	& 19.5 \\
        mag\_Deva & 17.4 & 16.0	& 21.3 \\
        mai\_Deva & 9.6	& 7.1 & 16.1 \\
        \midrule
        \textbf{Average} & 17.3 & 15.9 & 20.7 \\ 
        \bottomrule
    \end{tabular}}
    \caption{Using the NLLB dataset to noise the languages in the Indic family. The \mm{} model was tested. These results are computed over a 300 sample-subset of the FloRes-200 test set.}
    \label{nllb_noising}
\end{table}

\begin{table}[!ht]
    \centering
    \resizebox{0.4\textwidth}{!}{
    \begin{tabular}{l r r}
    \toprule
    \textbf{Language Family} & \textbf{\ttt{-cloud}} & \textbf{\ttt{-cloud-sem}} \\
    \midrule
    Austronesian & 17.9 & 18.0 \\
    Arabic & 22.8 & 22.6 \\
    Romance & 30.5 & 30.5 \\
    Turkic & 11.8 & 12.0 \\
    Indic & 21.0 & 21.0 \\
    French-Creole & 12.1 & - \\
    \bottomrule
    \end{tabular}
    }
    \caption{Additional semantic noisers as described in \autoref{app:choosing_noising_parameters} for \mtd{} are applied to the model \mm. Language family means were taken.}
    \label{tab:using_different_thetas}
\end{table}

\begin{table*}[htbp]
	\tiny
	\centering
	\resizebox{0.5\textwidth}{!}{
		\begin{tabular}{l | r | r r r}
			 \toprule
			 \textbf{Language} & \ttt{off-the-shelf} & \ttt{-cont} & \ttt{-func} & \ttt{-all}  \\
			 \midrule
\midrule
\multicolumn{4}{l}{\textbf{Austronesian Language Family}} \\
\midrule
ind\_Latn (Indonesian) & 38.9 & - & - & -  \\
\midrule
jav\_Latn (Javanese) & 22.9 & -0.2 & \textbf{\underline{+0.2}} & \textbf{\underline{+0.2}}  \\
sun\_Latn (Sundanese) & 22.9 & -2.6 & \textbf{\underline{+1.0}} & -1.5  \\
smo\_Latn (Samoan) & 2.3 & +0.2 & +1.0 & \textbf{\underline{+2.1}}  \\
mri\_Latn (Maori) & 2.3 & +0.5 & +1.2 & \textbf{\underline{+2.3}}  \\
ceb\_Latn (Cebuano) & 22.2 & \textbf{\underline{-2.7}} & -9.5 & -10.4  \\
\bottomrule
		 \end{tabular}
	 }
	\caption{\dtm{} BLEU scores by language for the model \mm. Performance gains/losses are relative to \ttt{off-the-shelf}. Averages for each \dtm{} approach are computed for each language family as well as for the general body of languages studied. These averages are recomputed for languages whose \ttt{off-the-shelf} BLEU score < 25. The best score for each language in the \dtm{} paradigm is \textbf{bolded} and \underline{underlined}.}
	\label{dtm_m2mL_bleu}
\end{table*}

\begin{table*}[htbp]
	\tiny
	\centering
	\resizebox{0.5\textwidth}{!}{
		\begin{tabular}{l | r | r r r}
			 \toprule
			 \textbf{Language} & \ttt{off-the-shelf} & \ttt{-cont} & \ttt{-func} & \ttt{-all}  \\
			 \midrule
\midrule
\multicolumn{4}{l}{\textbf{Austronesian Language Family}} \\
\midrule
ind\_Latn (Indonesian) & 41.5 & - & - & -  \\
\midrule
jav\_Latn (Javanese) & 14.2 & +2.9 & +7.8 & \textbf{\underline{+10.6}}  \\
sun\_Latn (Sundanese) & 11.0 & +3.6 & +10.1 & \textbf{\underline{+11.4}}  \\
smo\_Latn (Samoan) & 3.2 & -0.1 & +0.4 & \textbf{\underline{+1.1}}  \\
mri\_Latn (Maori) & 4.8 & \textbf{\underline{+0.4}} & -0.8 & \textbf{\underline{+0.4}}  \\
ceb\_Latn (Cebuano) & 14.0 & \textbf{\underline{-0.0}} & -3.0 & -1.7  \\
\bottomrule
		 \end{tabular}
	 }
	\caption{\dtm{} BLEU scores by language for the model \aya. Performance gains/losses are relative to \ttt{off-the-shelf}. Averages for each \dtm{} approach are computed for each language family as well as for the general body of languages studied. These averages are recomputed for languages whose \ttt{off-the-shelf} BLEU score < 25. The best score for each language in the \dtm{} paradigm is \textbf{bolded} and \underline{underlined}.}
	\label{dtm_aya-23-8b_bleu}
\end{table*}

\begin{table*}[htbp]
	\tiny
	\centering
	\resizebox{0.5\textwidth}{!}{
		\begin{tabular}{l | r | r r r}
			 \toprule
			 \textbf{Language} & \ttt{off-the-shelf} & \ttt{-cont} & \ttt{-func} & \ttt{-all}  \\
			 \midrule
\midrule
\multicolumn{5}{l}{\textbf{Austronesian Language Family}} \\
\midrule
ind\_Latn (Indonesian) & 87.4 & - & - & - \\
\midrule
jav\_Latn (Javanese) & 70.1 & -0.6 & -0.0 & -0.6 \\
sun\_Latn (Sundanese) & 71.6 & -2.6 & +0.9 & -2.7 \\
smo\_Latn (Samoan) & 27.4 & +7.5 & +3.9 & +13.6 \\
mri\_Latn (Maori) & 25.7 & +7.4 & +7.9 & +17.0 \\
ceb\_Latn (Cebuano) & 64.2 & -3.3 & -8.4 & -9.8 \\
\bottomrule
		 \end{tabular}
	 }
	\caption{\dtm{} COMET scores by language for the model \mm. Performance gains/losses are relative to \ttt{off-the-shelf}. Averages for each \dtm{} approach are computed for each language family as well as for the general body of languages studied. These averages are recomputed for languages whose \ttt{off-the-shelf} BLEU score < 25.}
	\label{dtm_m2mL_comet}
\end{table*}

\begin{table*}[htbp]
	\tiny
	\centering
	\resizebox{0.5\textwidth}{!}{
		\begin{tabular}{l | r | r r r}
			 \toprule
			 \textbf{Language} & \ttt{off-the-shelf} & \ttt{-cont} & \ttt{-func} & \ttt{-all}  \\
			 \midrule
\midrule
\multicolumn{5}{l}{\textbf{Austronesian Language Family}} \\
\midrule
ind\_Latn (Indonesian) & 88.5 & - & - & - \\
\midrule
jav\_Latn (Javanese) & 67.5 & +1.1 & +3.5 & +3.4 \\
sun\_Latn (Sundanese) & 63.7 & +1.7 & +7.6 & +6.1 \\
smo\_Latn (Samoan) & 35.4 & +3.9 & +0.5 & +5.3 \\
mri\_Latn (Maori) & 40.5 & +3.6 & -2.8 & +3.4 \\
ceb\_Latn (Cebuano) & 61.1 & -2.2 & -7.7 & -7.7 \\
\bottomrule
		 \end{tabular}
	 }
	\caption{\dtm{} COMET scores by language for the model \aya. Performance gains/losses are relative to \ttt{off-the-shelf}. Averages for each \dtm{} approach are computed for each language family as well as for the general body of languages studied. These averages are recomputed for languages whose \ttt{off-the-shelf} BLEU score < 25.}
	\label{dtm_aya-23-8b_comet}
\end{table*}

\begin{table*}[hbtp]
    \centering
    \resizebox{\textwidth}{!}{
    \begin{tabular}{l l l c c c c} 
        \toprule
        \textbf{Dictionary} & \textbf{Medium} & \textbf{Length (words)} & $\rightarrow$ \textbf{English} & $\rightarrow$ \textbf{Related Language} & \textbf{Inflections} & \textbf{Part-of-Speech} \\
        \midrule
        
        \multicolumn{7}{l}{\textbf{Fijian}} \\
        Ronald Gatty's Fijian-English Dictionary \citep{gatty_fijian} & PDF & $\sim$ 7585 & \checkmark & & & \checkmark \\
        Geocities Fijian-English Dictionary \tablefootnote{\url{https://web.archive.org/web/20091027151616/http://geocities.com/fijidictionary/eng_fijian_dict.html}} & table & $\sim$ 867 & \checkmark & & & \checkmark \\ %
        \midrule 
        
        \multicolumn{7}{l}{\textbf{Ilocano}} \\
        The Online Ilokano Dictionary Project \tablefootnote{\url{https://toidp.com/search/}} & mySQL database & $\sim$ 2060 & \checkmark & & & \\ %
        The Pinoy Dictionary \tablefootnote{\url{https://ilocano.pinoydictionary.com/}} & web app & unknown & \checkmark & & & \checkmark \\ %
        \midrule
        
        \multicolumn{7}{l}{\textbf{Samoan}} \\
\bottomrule
    \end{tabular}}
    \caption{Lexicons specific to a language}
    \label{tab:bilingual_lexicon_survey}
\end{table*}

\begin{table*}[hbtp]
	\tiny
	\centering
	\resizebox{0.9\textwidth}{!}{
	\begin{tabular}{l l l l l}
	\toprule
	\textbf{Dictionary Source} & \textbf{Total Words} & \textbf{Functional Words} & \textbf{Content Words} & \textbf{Coverage (\%))}\\
	\midrule
	\midrule
	\multicolumn{4}{c}{\textbf{French - Italian}} \\
	\midrule
		\textbf{Combined} & 238677 & 643 & 238034 & 89.3 \\
		PanLex \citep{kamholz-etal-2014-panlex} & 234647 & 370 & 234277 & 38.2 \\
		Swadesh \citep{swadesh} & 217 & 23 & 194 & 1.4 \\
		Statistically Aligned FloRes Data \citep{dyer2013simple} & 6636 & 538 & 6098 & 84.1 \\
	\midrule
	\multicolumn{4}{c}{\textbf{Spanish - Italian}} \\
	\midrule
		\textbf{Combined} & 209274 & 538 & 208736 & 90.7 \\
\end{tabular}
	}
	\caption{Part 1 \textemdash Total, functional, and content word counts across all lexicons. We also report the word type coverage, a measure of how extensively the lexicon documents a LRL. To calculate coverage percentage, we compared the CRL lexicon against the vocabulary size of the FloRes dev set.}
	\label{tab:bilingual_word_count_1}
\end{table*}

\begin{table*}[hbtp]
	\tiny
	\centering
	\resizebox{0.9\textwidth}{!}{
	\begin{tabular}{l l l l l}
	\toprule
	\textbf{Dictionary Source} & \textbf{Total Words} & \textbf{Functional Words} & \textbf{Content Words} & \textbf{Coverage (\%))}\\
	\midrule
	\midrule
	\multicolumn{4}{c}{\textbf{Sicilian - Italian}} \\
	\midrule
		\textbf{Combined} & 23755 & 575 & 23180 & 90.2 \\
		Dieli \citep{dieli} & 10380 & 143 & 10237 & 11.8 \\
		PanLex \citep{kamholz-etal-2014-panlex} & 17080 & 206 & 16874 & 15.9 \\
		Swadesh \citep{swadesh} & 219 & 22 & 197 & 1.2 \\
		Statistically Aligned FloRes Data \citep{dyer2013simple} & 7213 & 528 & 6685 & 88.8 \\
	\midrule
	\multicolumn{4}{c}{\textbf{Northern Uzbek - Turkish}} \\
\end{tabular}
	}
	\caption{Part 2 \textemdash Total, functional, and content word counts across all lexicons. We also report the word type coverage, a measure of how extensively the lexicon documents a LRL. To calculate coverage percentage, we compared the CRL lexicon against the vocabulary size of the FloRes dev set.}
	\label{tab:bilingual_word_count_2}
\end{table*}

\begin{table*}[hbtp]
	\tiny
	\centering
	\resizebox{0.95\textwidth}{!}{
	\begin{tabular}{l l l l l}
	\toprule
	\textbf{Dictionary Source} & \textbf{Total Words} & \textbf{Functional Words} & \textbf{Content Words} & \textbf{Coverage (\%))}\\
	\midrule
	\midrule
	\multicolumn{4}{c}{\textbf{Awadhi - Hindi}} \\
	\midrule
		\textbf{Combined} & 10470 & 1356 & 9114 & 31.8 \\
		PanLex \citep{kamholz-etal-2014-panlex} & 12 & 0 & 12 & 0.0 \\
		\citep{bafna-etal-2024-cousin} & 10462 & 1355 & 9107 & 31.8 \\
	\midrule
	\multicolumn{4}{c}{\textbf{Tagalog - Indonesian}} \\
	\midrule
		\textbf{Combined} & 10176 & 284 & 9892 & 79.5 \\
		PanLex \citep{kamholz-etal-2014-panlex} & 4916 & 76 & 4840 & 8.9 \\
\end{tabular}
	}
	\caption{Part 3 \textemdash Total, functional, and content word counts across all lexicons. We also report the word type coverage, a measure of how extensively the lexicon documents a LRL. To calculate coverage percentage, we compared the CRL lexicon against the vocabulary size of the FloRes dev set.}
	\label{tab:bilingual_word_count_3}
\end{table*}

\begin{table*}[hbtp]
	\tiny
	\centering
	\resizebox{0.95\textwidth}{!}{
	\begin{tabular}{l l l l l}
	\toprule
	\textbf{Dictionary Source} & \textbf{Total Words} & \textbf{Functional Words} & \textbf{Content Words} & \textbf{Coverage (\%))}\\
	\midrule
	\midrule
	\multicolumn{4}{c}{\textbf{North Levantine Arabic - Standard Arabic}} \\
	\midrule
		Statistically Aligned FloRes Data \citep{dyer2013simple} & 8411 & 416 & 7995 & 93.6 \\
	\midrule
	\multicolumn{4}{c}{\textbf{Italian - Italian}} \\
	\midrule
		Statistically Aligned FloRes Data \citep{dyer2013simple} & 8118 & 226 & 7892 & 98.3 \\
	\midrule
	\multicolumn{4}{c}{\textbf{Plateau Malagasy - Indonesian}} \\
	\midrule
		\textbf{Combined} & 8106 & 433 & 7673 & 85.9 \\
		PanLex \citep{kamholz-etal-2014-panlex} & 2420 & 85 & 2335 & 3.5 \\
\end{tabular}
	}
	\caption{Part 4 \textemdash Total, functional, and content word counts across all lexicons. We also report the word type coverage, a measure of how extensively the lexicon documents a LRL. To calculate coverage percentage, we compared the CRL lexicon against the vocabulary size of the FloRes dev set.}
	\label{tab:bilingual_word_count_4}
\end{table*}

\begin{table*}[hbtp]
	\centering
	\resizebox{\textwidth}{!}{
	\begin{tabular}{l l l l l}
	\midrule
	\textbf{Dictionary Source} & \textbf{Total Words} & \textbf{Functional Words} & \textbf{Content Words} & \textbf{Coverage (\%)} \\
	\midrule
	\midrule
	\multicolumn{4}{c}{\textbf{Fijian - Indonesian}} \\
	\midrule
		\textbf{Combined} & 4280 & 205 & 4075 & 77.5 \\
		PanLex \citep{kamholz-etal-2014-panlex} & 630 & 46 & 584 & 4.1 \\
		Swadesh \citep{swadesh} & 222 & 33 & 189 & 1.3 \\
		Statistically Aligned FloRes Data \citep{dyer2013simple} & 3705 & 150 & 3555 & 77.0 \\
		\bottomrule
	\end{tabular}
	}
	\caption{Part 5 \textemdash Total, functional, and content word counts across all lexicons. We also report the word type coverage, a measure of how extensively the lexicon documents a LRL. To calculate coverage percentage, we compared the CRL lexicon against the vocabulary size of the FloRes dev set.}
	\label{tab:bilingual_word_count_5}
\end{table*}

\begin{table}
\small
\begin{tabular}{lrrrr}
\toprule
 & \multicolumn{2}{c}{Aya} & \multicolumn{2}{c}{M2M} \\
 \midrule
feature                 & \ttt{lrc}        & \ttt{fi}             & \ttt{lrc}        & \ttt{fi}               \\
\midrule
\ttt{lrl\_wiki\_count}        & 0.0           & 0.07    & 0.0           & 0.02    \\
\ttt{hrl\_wiki\_count}        & 0.0            & 0.11    & 0.0            & 0.2     \\
\ttt{chrf\_sim}               & -0.03          & 0.12    & -0.07          & 0.26    \\
\ttt{baseline\_performance}   & -0.3           & 0.48    & -0.12          & 0.45    \\
\ttt{in\_pretraining}         & -1.72          & 0.0     & -1.67          & 0.0     \\
\ttt{token\_fertility\_ratio} & -6.3           & 0.21    & -3.66          & 0.07   \\
\bottomrule
\end{tabular}
\caption{Random forest feature importance values and linear regression coefficients for different language features. \ttt{lrc} = linear regression coefficient; \ttt{fi} = feature importance value.}
\label{feat_imps}
\end{table}

\begin{table}[ht]
    \centering
    \tiny
    \resizebox{0.45\textwidth}{!}{
    \begin{tabular}{l r r}
    
        \toprule
        \textbf{HRL} & \textbf{Natural \%} & \textbf{Mean \% over all CRLs} \\
        \midrule
        ind & 24.5 & 43.1 \\
        arb & 27.7 & 16.3 \\
        ita & 40.3 & 38.5 \\
        tur & 16.8 & 15.0 \\
        hin & 39.7 & 33.4 \\
        \bottomrule
    \end{tabular}
    }
    \caption{Comparing the natural percentage of functional words in an HRL against the mean \% of words identified as function words over all its CRLs for \dtm{}.}
    \label{natural_vs_changed_distribution}
\end{table}

\begin{table*}[htbp]
	\tiny
	\centering
	\resizebox{0.4\textwidth}{!}{
		\begin{tabular}{l | r r r}
			 \toprule
			 \textbf{Language} & \textbf{\% in \ttt{-cont}} & \textbf{\% in \ttt{-func}} & \textbf{\% in \ttt{-all}} \\
			 \midrule
\midrule
\multicolumn{4}{l}{\textbf{Austronesian Language Family}} \\
\midrule
jav\_Latn (Javanese) & 42.9 & 32.2 & 75.1 \\
sun\_Latn (Sundanese) & 40.2 & 31.7 & 71.9 \\
smo\_Latn (Samoan) & 33.2 & 52.2 & 85.4 \\
mri\_Latn (Maori) & 41.2 & 46.8 & 87.9 \\
ceb\_Latn (Cebuano) & 26.0 & 49.7 & 75.7 \\
zsm\_Latn (Standard Malay) & 61.8 & 27.5 & 89.3 \\
\bottomrule
		 \end{tabular}
	 }
	\caption{Percentage of words switched out in \dtm {} approaches for each language}
	\label{changed_words_across_langs}
\end{table*}

\begin{table}[hbtp]
	\resizebox{0.48\textwidth}{!}{
	\begin{tabular}{l l l l l}
		\hline
			& \textbf{\ttt{off-the-shelf}} & \textbf{\ttt{-all-eng}} & \textbf{\ttt{-cont-eng}} & \textbf{\ttt{-func-eng}} \\
		\hline
			ind (Indonesian) & 15.9 & 10.1 & 10.2 & 13.4 \\
			arb (Standard Arabic) & 29.5 & 10.3 & 10.2 & 28.8 \\
			ita (Italian) & 29.3 & 14.4 & 16.6 & 22.8 \\
			tur (Turkish) & 12.5 & 3.7 & 4.1 & 7.6 \\
			hin (Hindi) & 22.9 & 4.9 & 4.8 & 19.5 \\
		\hline
	\end{tabular}
	}
	\caption{\dtm {} results for \aya {} where low-resource words are swapped with English words. Experiment was performed on 300 FLORES examples.}
	\label{dtm_aya-23-8b_eng}
\end{table}

\begin{table}[hbtp]
	\resizebox{0.48\textwidth}{!}{
	\begin{tabular}{l l l l l}
		\hline
			& \textbf{\ttt{off-the-shelf}} & \textbf{\ttt{-all-eng}} & \textbf{\ttt{-cont-eng}} & \textbf{\ttt{-func-eng}} \\
		\hline
			ind (Indonesian) & 13.5 & 11.1 & 11.0 & 11.0 \\
			arb (Standard Arabic) & 20.9 & 9.3 & 9.3 & 19.9 \\
			ita (Italian) & 19.6 & 14.3 & 14.3 & 16.8 \\
			tur (Turkish) & 10.9 & 4.6 & 4.7 & 5.5 \\
			hin (Hindi) & 17.3 & 5.4 & 5.4 & 12.9 \\
		\hline
	\end{tabular}
	}
	\caption{\dtm {} results for \mm {} where low-resource words are swapped with English words  Experiment was performed on 300 FLORES examples.}
	\label{dtm_m2mL_eng}
\end{table}

\begin{table*}[htbp]
	\tiny
	\centering
	\resizebox{0.6\textwidth}{!}{
		\begin{tabular}{l | r | r r r}
			 \toprule
			 \textbf{Language} & \ttt{off-the-shelf} & \ttt{-cloud-cont} & \ttt{-cloud-func} & \ttt{-cloud-all}  \\
			 \midrule
\midrule
\multicolumn{4}{l}{\textbf{Austronesian Language Family}} \\
\midrule
ind\_Latn (Indonesian) & 38.9 & - & - & -  \\
\midrule
jav\_Latn (Javanese) & 22.9 & +0.4 & \textbf{\underline{+1.1}} & +0.7  \\
sun\_Latn (Sundanese) & 22.9 & -1.6 & \textbf{\underline{+1.5}} & -1.2  \\
smo\_Latn (Samoan) & 2.3 & +1.1 & +1.6 & \textbf{\underline{+2.8}}  \\
mri\_Latn (Maori) & 2.3 & +2.4 & +1.4 & \textbf{\underline{+2.8}}  \\
ceb\_Latn (Cebuano) & 22.2 & \textbf{\underline{-1.9}} & -7.3 & -9.1  \\
\bottomrule
		 \end{tabular}
	 }
	\caption{\combined{} BLEU scores by language for the model \mm. Performance gains/losses are relative to \ttt{off-the-shelf}. Averages for each \combined{} approach are computed for each language family as well as for the general body of languages studied. These averages are recomputed for languages whose \ttt{off-the-shelf} BLEU score < 25. The best score for each language in the \combined{} paradigm is \textbf{bolded} and \underline{underlined}.}
	\label{mtd+dtm_m2mL_bleu}
\end{table*}

\begin{table*}[htbp]
	\tiny
	\centering
	\resizebox{0.6\textwidth}{!}{
		\begin{tabular}{l | r | r r r}
			 \toprule
			 \textbf{Language} & \ttt{off-the-shelf} & \ttt{-cloud-cont} & \ttt{-cloud-func} & \ttt{-cloud-all}  \\
			 \midrule
\midrule
\multicolumn{4}{l}{\textbf{Austronesian Language Family}} \\
\midrule
ind\_Latn (Indonesian) & 41.5 & - & - & -  \\
\midrule
jav\_Latn (Javanese) & 14.2 & +7.6 & +9.9 & \textbf{\underline{+12.2}}  \\
sun\_Latn (Sundanese) & 11.0 & +8.5 & +12.3 & \textbf{\underline{+13.3}}  \\
smo\_Latn (Samoan) & 3.2 & +1.8 & +1.9 & \textbf{\underline{+2.2}}  \\
mri\_Latn (Maori) & 4.8 & \textbf{\underline{+2.0}} & -0.0 & +1.8  \\
ceb\_Latn (Cebuano) & 14.0 & \textbf{\underline{-0.3}} & -2.1 & -0.7  \\
\bottomrule
		 \end{tabular}
	 }
	\caption{\combined{} BLEU scores by language for the model \aya. Performance gains/losses are relative to \ttt{off-the-shelf}. Averages for each \combined{} approach are computed for each language family as well as for the general body of languages studied. These averages are recomputed for languages whose \ttt{off-the-shelf} BLEU score < 25. The best score for each language in the \combined{} paradigm is \textbf{bolded} and \underline{underlined}.}
	\label{mtd+dtm_aya-23-8b_bleu}
\end{table*}

\begin{table*}[htbp]
	\tiny
	\centering
	\resizebox{0.6\textwidth}{!}{
		\begin{tabular}{l | r | r r r}
			 \toprule
			 \textbf{Language} & \ttt{off-the-shelf} & \ttt{-cloud-cont} & \ttt{-cloud-func} & \ttt{-cloud-all}  \\
			 \midrule
\midrule
\multicolumn{5}{l}{\textbf{Austronesian Language Family}} \\
\midrule
ind\_Latn (Indonesian) & 87.4 & - & - & - \\
\midrule
jav\_Latn (Javanese) & 70.1 & +0.4 & +1.3 & +0.1 \\
sun\_Latn (Sundanese) & 71.6 & -1.7 & +1.5 & -1.9 \\
smo\_Latn (Samoan) & 27.4 & +12.4 & +7.5 & +15.5 \\
mri\_Latn (Maori) & 25.7 & +16.6 & +10.4 & +18.7 \\
ceb\_Latn (Cebuano) & 64.2 & -2.0 & -5.6 & -7.9 \\
\bottomrule
		 \end{tabular}
	 }
	\caption{\combined{} COMET scores by language for the model \mm. Performance gains/losses are relative to \ttt{off-the-shelf}. Averages for each \combined{} approach are computed for each language family as well as for the general body of languages studied. These averages are recomputed for languages whose \ttt{off-the-shelf} BLEU score < 25.}
	\label{mtd+dtm_m2mL_comet}
\end{table*}

\begin{table*}[htbp]
	\tiny
	\centering
	\resizebox{0.7\textwidth}{!}{
		\begin{tabular}{l | r | r r r}
			 \toprule
			 \textbf{Language} & \ttt{off-the-shelf} & \ttt{-cloud-cont} & \ttt{-cloud-func} & \ttt{-cloud-all}  \\
			 \midrule
\midrule
\multicolumn{5}{l}{\textbf{Austronesian Language Family}} \\
\midrule
ind\_Latn (Indonesian) & 88.5 & - & - & - \\
\midrule
jav\_Latn (Javanese) & 67.5 & +6.6 & +8.2 & +8.0 \\
sun\_Latn (Sundanese) & 63.7 & +8.9 & +12.1 & +11.1 \\
smo\_Latn (Samoan) & 35.4 & +11.4 & +6.8 & +10.5 \\
mri\_Latn (Maori) & 40.5 & +9.0 & +0.9 & +8.0 \\
ceb\_Latn (Cebuano) & 61.1 & +0.3 & -3.8 & -3.0 \\
\bottomrule
		 \end{tabular}
	 }
	\caption{\combined{} COMET scores by language for the model \aya. Performance gains/losses are relative to \ttt{off-the-shelf}. Averages for each \combined{} approach are computed for each language family as well as for the general body of languages studied. These averages are recomputed for languages whose \ttt{off-the-shelf} BLEU score < 25.}
	\label{mtd+dtm_aya-23-8b_comet}
\end{table*}

\end{document}